%% file: acl_latex.tex
\pdfoutput=1

\documentclass[11pt]{article}

\usepackage[preprint]{acl}

\usepackage{times}
\usepackage{latexsym}

\usepackage[T1]{fontenc}

\usepackage[utf8]{inputenc}

\usepackage{microtype}

\usepackage{inconsolata}

\usepackage{graphicx}
\usepackage{hyperref}

\usepackage{todonotes}
\usepackage{multirow}
\usepackage{amsmath,amssymb,amsfonts}
\usepackage{amsthm}
\usepackage{mathrsfs}
\usepackage{xcolor}
\usepackage{textcomp}
\usepackage{manyfoot}
\usepackage{booktabs}
\usepackage{algorithm}
\usepackage{algorithmicx}
\usepackage{algpseudocode}
\usepackage{listings}
\usepackage{hyperref}       
\usepackage{url}            
\usepackage{nicefrac}       
\usepackage{microtype}      
\usepackage[capitalise,nameinlink]{cleveref}
\usepackage{caption}
\usepackage[acronym,nolist,nogroupskip]{glossaries}
\usepackage{siunitx}
\usepackage{wrapfig}        
\usepackage[font={small}]{caption}
\usepackage{comment}
\usepackage{framed}
\usepackage{xltabular}
\usepackage{chngcntr}
\usepackage{tcolorbox}
\usepackage{fancyvrb}
\usepackage{fvextra}
\usepackage{placeins}
\usepackage{afterpage}
\usepackage{subcaption}
\usepackage{paralist}
\usepackage{longtable}
\usepackage{tabularx}
\usepackage{ltablex}
\usepackage{changepage}
\usepackage{aliascnt}

\newacronym{SFSC}{SFSC}{Swiss Federal Supreme Court}
\newacronym{SCI}{SCI}{Supreme Court of India}
\newacronym{ECHR}{ECHR}{European Convention of Human Rights}
\newacronym{ECTHR}{ECtHR}{European Court of Human Rights}
\newacronym{SCOTUS}{SCOTUS}{Supreme Court of the United States}
\newacronym{SPC}{SPC}{Supreme People's Court of China}
\newacronym{SJP}{SJP}{Swiss-Judgment-Prediction}
\newacronym{ASO}{ASO}{Almost Stochastic Order}
\newacronym{ILDC}{ILDC}{Indian Legal Documents Corpus}

\newacronym{US}{US}{United States}
\newacronym{EU}{EU}{European Union}

\newacronym{NLP}{NLP}{Natural Language Processing}
\newacronym{ML}{ML}{Machine Learning}
\newacronym{LJP}{LJP}{Legal Judgment Prediction}

\newacronym{BERT}{BERT}{Bidirectional Encoder Representations from Transformers}

\newacronym{PLM}{PLM}{Pre-trained Language Model}

\newacronym{CLT}{CLT}{Cross-Lingual Transfer}
\newacronym{LD}{LD}{Landmark Decision}
\newacronym{FSCD}{FSCD}{Federal Supreme Court Decisions}
\newacronym{SCPC}{SCPC}{Swiss Civil Procedure Code}
\newacronym{LLM}{LLM}{Large Language Model}
\newacronym{SLM}{SLM}{Small Language Model}
\newacronym{LM}{LM}{Language Model}
\newacronym{PEFT}{PEFT}{Parameter Efficient Fine-Tuning}

\newacronym{CR}{CR}{Compression Ratio}
\newacronym{EFC}{EFC}{Extractive Fragment Coverage}
\newacronym{EFD}{EFD}{Extractive Fragment Density}

\newacronym{SLDS}{SLDS}{Swiss Landmark Decision Summarization}

\newacronym{CI}{CI}{Coverage Increment}

\glsdisablehyper

\title{Unlocking Legal Knowledge:\\A Multilingual Dataset for Judicial Summarization in Switzerland}

\author{
Luca Rolshoven $^{1,2}$
\And
Vishvaksenan Rasiah $^{1}$
\And
Srinanda Brügger Bose $^{4}$
\AND
Sarah Hostettler $^{1,2}$
\And
Lara Burkhalter $^{1,2}$
\And
Matthias Stürmer $^{1,2}$
\And
Joel Niklaus $^{1,2,3}$
\AND
\\
$^1$University of Bern\quad
$^2$Bern University of Applied Sciences\\
$^3$Stanford University
$^4$University of Fribourg\quad
}

\DeclareCaptionType{prompt}[Prompt][List of Prompts]
\DeclareCaptionType{example}[Example][List of Examples]

\input{environments_and_colors}

\crefalias{prompt}{figure}
\crefname{prompt}{prompt}{prompts}
\Crefname{prompt}{Prompt}{Prompts}
\crefalias{example}{figure}
\crefname{example}{example}{examples}
\Crefname{example}{Example}{Examples}

\begin{document}
\maketitle

\begin{abstract}
    \input{sections/abstract}
\end{abstract}

\input{sections/introduction}
\input{sections/related_work}
\input{sections/dataset}
\input{sections/setup}
\input{sections/results}

\input{sections/expert_evaluation}
\input{sections/conclusions}
\input{sections/limitations}
\input{sections/acknowledgements}

\bibliography{custom}

\clearpage

\appendix

\input{sections/appendix}

\end{document}

%% file: environments_and_colors.tex
\definecolor{metadataColor}{HTML}{2F416B} 
\definecolor{inputColor}{HTML}{6B4B2F}
\definecolor{outputColor}{HTML}{6B2F48}
\definecolor{targetColor}{HTML}{556B2F}
\definecolor{evaluationsColor}{RGB}{105, 105, 105}
\definecolor{tablePalette1}{HTML}{fa8775}
\definecolor{tablePalette2}{HTML}{ea5f94}
\definecolor{tablePalette3}{HTML}{cd34b5}
\definecolor{tablePalette4}{HTML}{9d02d7}
\definecolor{tablePalette5}{HTML}{0000ff}

\colorlet{containerFrame}{black}
\colorlet{containerBg}{black!5!white}
\colorlet{metadataFrame}{metadataColor}
\colorlet{metadataBg}{metadataColor!10!white}
\colorlet{inputFrame}{inputColor}
\colorlet{inputBg}{inputColor!10!white}
\colorlet{outputFrame}{outputColor}
\colorlet{outputBg}{outputColor!10!white}
\colorlet{targetFrame}{targetColor}
\colorlet{targetBg}{targetColor!10!white}
\colorlet{metricsFrame}{evaluationsColor}
\colorlet{metricsBg}{evaluationsColor!10!white}

{%
    \begin{tcolorbox}[title=#1, width=1.0\textwidth, top=1ex, bottom=1ex, left=1ex, right=1ex, colframe=containerFrame, colback=containerBg, valign=center, halign=left]
        \begin{tcolorbox}[title=Target, width=1.0\textwidth, top=1ex, bottom=1ex, left=1ex, right=1ex, colframe=targetFrame, colback=targetBg, valign=center, halign=left]
        #2
        \end{tcolorbox}
        \begin{tcolorbox}[title=Predicted, width=1.0\textwidth, top=1ex, bottom=1ex, left=1ex, right=1ex, colframe=outputFrame, colback=outputBg, valign=center, halign=left]
        #3
        \end{tcolorbox}
        \begin{tcolorbox}[title=Metrics, width=1.0\textwidth, top=1ex, bottom=1ex, left=1ex, right=1ex, colframe=metricsFrame, colback=metricsBg, valign=center, halign=left]
        #4
        \end{tcolorbox}
    }
{%
    \end{tcolorbox}
}

{%
    \begin{tcolorbox}[title=\small{Reference Sample}, width=1.0\textwidth, top=1ex, bottom=1ex, left=1ex, right=1ex, colframe=containerFrame, colback=containerBg, valign=center, halign=left]
        \begin{tcolorbox}[title=\small{Decision (Input)}, width=1.0\textwidth, top=1ex, bottom=1ex, left=1ex, right=1ex, colframe=inputFrame, colback=inputBg, valign=center, halign=left, sidebyside, sidebyside align=top]
            \scriptsize
            \textbf{Original:} #2
            \tcblower
            \scriptsize
            \textbf{Translated:} #3
        \end{tcolorbox}
        \begin{tcolorbox}[title=\small{Headnote (Ground Truth)}, width=1.0\textwidth, top=1ex, bottom=1ex, left=1ex, right=1ex, colframe=targetFrame, colback=targetBg, valign=center, halign=left, sidebyside, sidebyside align=top]
            \scriptsize
            \textbf{Original:} #4
            \tcblower
            \scriptsize
            \textbf{Translated:} #5
        \end{tcolorbox}
        \begin{tcolorbox}[title=\small{Metadata}, width=1.0\textwidth, top=1ex, bottom=1ex, left=1ex, right=1ex, colframe=metadataFrame, colback=metadataBg, valign=center, halign=left]
            \scriptsize
            #1
        \end{tcolorbox}
    }
{%
    \end{tcolorbox}
}

{%
    \begin{tcolorbox}[title=#1, width=1.0\textwidth, top=1ex, bottom=1ex, left=1ex, right=1ex, colframe=containerFrame, colback=containerBg, valign=center, halign=left]
        \begin{tcolorbox}[title=Headnote (Output), width=1.0\textwidth, top=1ex, bottom=1ex, left=1ex, right=1ex, colframe=outputFrame, colback=outputBg, valign=top, halign=left, sidebyside, sidebyside align=top]
            \small
            \textbf{Original:} #2
            \tcblower
            \small
            \textbf{Translated:} #3
        \end{tcolorbox}
        \begin{tcolorbox}[title=Evaluation, width=1.0\textwidth, top=1ex, bottom=1ex, left=1ex, right=1ex, colframe=metricsFrame, colback=metricsBg, valign=center, halign=left]
            #4
        \end{tcolorbox}
    }
{%
    \end{tcolorbox}
}

\newenvironment{judgeExample}[8] 
{%
    \noindent
    \makebox[\linewidth][c]{ 
        \begin{tcolorbox}[
            title=\small{LLM-as-a-Judge Output}, 
            width=\dimexpr\linewidth+2cm\relax, 
            top=1ex, bottom=1ex, left=1ex, right=1ex, 
            colframe=containerFrame, colback=containerBg, 
            valign=center, halign=left
        ]
            \begin{tcolorbox}[
                title=\small{Generated Headnote \hfill(Model: #2)}, 
                width=\dimexpr\linewidth\relax, 
                top=1ex, bottom=1ex, left=1ex, right=1ex, 
                colframe=inputFrame, colback=inputBg, 
                valign=center, halign=left, 
                sidebyside, sidebyside align=top
            ]
                \tiny
                \textbf{Original:} #4
                \tcblower
                \tiny
                \textbf{Translated:} #5
            \end{tcolorbox}

            \begin{tcolorbox}[
                title=\small{Original Headnote \hfill(Sample ID: #1)}, 
                width=\dimexpr\linewidth\relax, 
                top=1ex, bottom=1ex, left=1ex, right=1ex, 
                colframe=targetFrame, colback=targetBg, 
                valign=center, halign=left, 
                sidebyside, sidebyside align=top
            ]
                \tiny
                \textbf{Original:} #6
                \tcblower
                \tiny
                \textbf{Translated:} #7
            \end{tcolorbox}

            \begin{tcolorbox}[
                title=\small{Judge Output \hfill(Aggregated Score: #3)}, 
                width=\dimexpr\linewidth\relax, 
                top=1ex, bottom=1ex, left=1ex, right=1ex, 
                colframe=outputFrame, colback=outputBg, 
                valign=center, halign=left
            ]
                \tiny
                #8
            \end{tcolorbox}
        \end{tcolorbox}
    }
}
{}

\newenvironment{contextualizedEvaluation}[8] 
{%
    \noindent
    \makebox[\linewidth][c]{ 
        \begin{tcolorbox}[
            title=\small{Contextualized Human Evaluation (Sample #1)}, 
            width=\dimexpr\linewidth+2cm\relax, 
            top=1ex, bottom=1ex, left=1ex, right=1ex, 
            colframe=containerFrame, colback=containerBg, 
            valign=center, halign=left
        ]
            \begin{tcolorbox}[
                title=\small{Original Headnote\hfill\textit{(translated to English)}}, 
                width=\dimexpr\linewidth\relax, 
                top=1ex, bottom=1ex, left=1ex, right=1ex, 
                colframe=containerFrame, colback=containerBg, 
                valign=center, halign=left
            ]
                \tiny
                #2
            \end{tcolorbox}

            \begin{tcolorbox}[
                title=\small{Generated Headnote (Claude 3.5 Sonnet)\hfill\textit{(translated to English)}}, 
                width=\dimexpr\linewidth\relax, 
                top=1ex, bottom=1ex, left=1ex, right=1ex, 
                colframe=targetFrame, colback=targetBg, 
                valign=center, halign=left, 
                sidebyside, sidebyside align=top
            ]
                \tiny
                \textbf{Headnote:} #3
                \tcblower
                \tiny
                \textbf{Expert Commentary:} #4
            \end{tcolorbox}

            \begin{tcolorbox}[
                title=\small{Generated Headnote (DeepSeek R1)\hfill\textit{(translated to English)}}, 
                width=\dimexpr\linewidth\relax, 
                top=1ex, bottom=1ex, left=1ex, right=1ex, 
                colframe=inputFrame, colback=inputBg, 
                valign=center, halign=left, 
                sidebyside, sidebyside align=top
            ]
                \tiny
                \textbf{Headnote:} #5
                \tcblower
                \tiny
                \textbf{Expert Commentary:} #6
            \end{tcolorbox}

            \begin{tcolorbox}[
                title=\small{Generated Headnote (Qwen2.5 14B)\hfill\textit{(translated to English)}}, 
                width=\dimexpr\linewidth\relax, 
                top=1ex, bottom=1ex, left=1ex, right=1ex, 
                colframe=outputFrame, colback=outputBg, 
                valign=center, halign=left, 
                sidebyside, sidebyside align=top
            ]
                \tiny
                \textbf{Headnote:} #7
                \tcblower
                \tiny
                \textbf{Expert Commentary:} #8
            \end{tcolorbox}
        \end{tcolorbox}
    }
}
{}

\newenvironment{judgeExampleSmall}[7] 
{%
    \noindent
    \makebox[\linewidth][c]{ 
        \begin{tcolorbox}[
            title=\small{LLM-as-a-Judge Output}, 
            width=0.93\linewidth, 
            top=0.5ex, bottom=0.5ex, left=0.5ex, right=0.5ex, 
            colframe=containerFrame, colback=containerBg, 
            valign=center, halign=left
        ]
            \begin{tcolorbox}[
                title=\small{Decision \hfill(Sample ID: #1)}, 
                width=\dimexpr\linewidth\relax, 
                top=1ex, bottom=1ex, left=1ex, right=1ex, 
                colframe=containerFrame, colback=containerBg, 
                valign=center, halign=left
            ]
                \tiny
                #3
            \end{tcolorbox}
            
            \begin{tcolorbox}[
                title=\small{Original Headnote \hfill(Sample ID: #1)}, 
                width=\dimexpr\linewidth\relax, 
                top=1ex, bottom=1ex, left=1ex, right=1ex, 
                colframe=targetFrame, colback=targetBg, 
                valign=center, halign=left
            ]
                \tiny
                #4
            \end{tcolorbox}

            \begin{tcolorbox}[
                title=\small{Generated Headnote \hfill(Model: #2)}, 
                width=\dimexpr\linewidth\relax, 
                top=1ex, bottom=1ex, left=1ex, right=1ex, 
                colframe=inputFrame, colback=inputBg, 
                valign=center, halign=left
            ]
                \tiny
                #5
            \end{tcolorbox}

            \begin{tcolorbox}[
                title=\small{Judge Output \hfill(Aggregated Score: #6)}, 
                width=\dimexpr\linewidth\relax, 
                top=1ex, bottom=1ex, left=1ex, right=1ex, 
                colframe=outputFrame, colback=outputBg, 
                valign=center, halign=left
            ]
                \tiny
                #7
            \end{tcolorbox}
        \end{tcolorbox}
    }
}
{}

%% file: sections/abstract.tex
Legal research depends on headnotes: concise summaries that help lawyers quickly identify relevant cases. Yet, many court decisions lack them due to the high cost of manual annotation. To address this gap, we introduce the Swiss Landmark Decisions Summarization (SLDS) dataset containing 20K rulings from the Swiss Federal Supreme Court, each with headnotes in German, French, and Italian. SLDS has the potential to significantly improve access to legal information and transform legal research in Switzerland. We fine-tune open models (Qwen2.5, Llama 3.2, Phi-3.5) and compare them to larger general-purpose and reasoning-tuned LLMs, including GPT-4o, Claude 3.5 Sonnet, and the open-source DeepSeek R1. Using an LLM-as-a-Judge framework, we find that fine-tuned models perform well in terms of lexical similarity, while larger models generate more legally accurate and coherent summaries. Interestingly, reasoning-focused models show no consistent benefit, suggesting that factual precision is more important than deep reasoning in this task. We release SLDS under a CC BY 4.0 license to support future research in cross-lingual legal summarization.

%% file: sections/introduction.tex
\section{Introduction}

A significant part of legal work involves research, where lawyers must find similar cases and navigate numerous judicial decisions, especially when interpreting laws with room for debate. Due to the time-intensive nature of this task, they usually rely on judgment summaries. However, creating these summaries is labor intensive and requires the expertise of judges and clerks, who are already burdened with a large caseload \citep{bieri2015law} and time pressure \citep{ludewig2013professional}.

To alleviate this increasing need for efficient ways to navigate large amounts of legal texts, legal document summarization has become a critical area of interest in NLP \citep{JAIN2021100388}. Over the years, researchers have made significant strides in both extractive and abstractive summarization of legal texts. Earlier work focused on extracting key sentences to create concise summaries \citep{grover_holj_2004, hachey-grover-2006-extractive, 10.1007/978-3-642-39931-2_14, bhattacharya2021incorporating}, while recent advancements have turned toward abstractive methods, which generate condensed paraphrases of the most important information in a document \citep{shukla2022legal, niklaus_budgetlongformer_2022, moro2023multi, jain2024summarization, niklaus_flawn-t5_2024}.

\begin{figure}[t]
    \centering
    \includegraphics[width=1\linewidth]{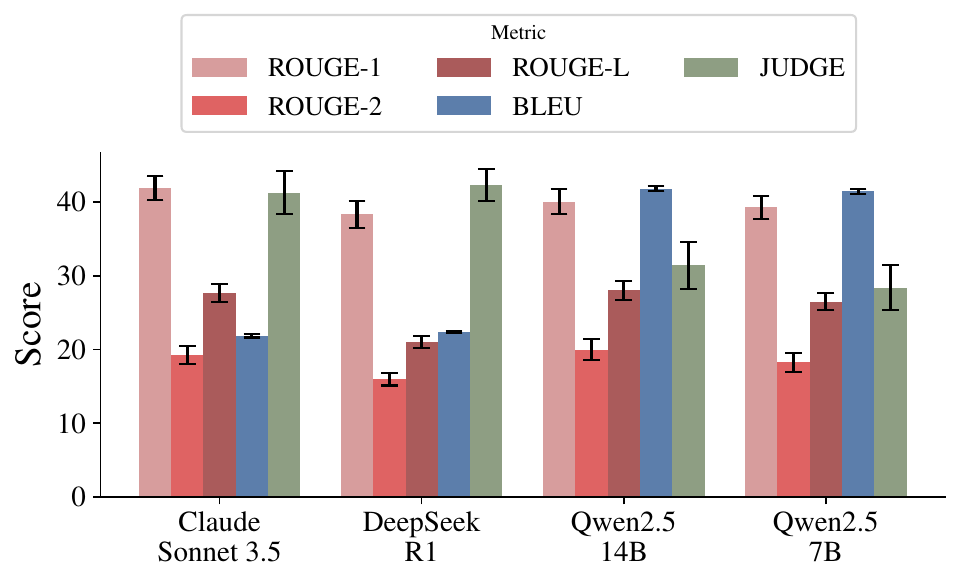}
    \vspace{-4ex}
    \caption{Two fine-tuned LLMs of the Qwen2.5 family and two frontier models evaluated on the SLDS test set. While fine-tuning dominates outcomes in terms of lexical metrics, the smaller fine-tuned models do not yet reach the same output quality as their larger pre-trained counterparts, as indicated by the LLM-as-a-Judge \citep{zheng2023judging} score.}
    \label{fig:results-bar-chart}
    \vspace{-3.5ex}
\end{figure}

Datasets with legal documents and their corresponding summaries have been instrumental in enabling these advancements, yet they primarily focus on monolingual corpora or multiple jurisdictions. Therefore, existing datasets do not adequately address the unique challenges posed by multilingual jurisdictions, such as Switzerland, where legal decisions are written in multiple languages and need to be summarized consistently. This gap is particularly relevant because many legal NLP tools and models are trained on English-centric datasets, which may not reliably generalize to cross-lingual environments.

We introduce \gls{SLDS}, a large-scale multilingual dataset of Swiss Supreme Court cases in German, French, and Italian, featuring headnotes that summarize key legal points and laws. By focusing on these concise legal digests, SLDS facilitates cross-lingual legal summarization research and supports the development of tools for professionals working across language barriers. The dataset is publicly available under a CC BY 4.0 license.\footnote{Available at \url{https://huggingface.co/datasets/ipst/slds}}

\paragraph{Contributions} Our contributions are two-fold:

\begin{compactenum}
    \item \textbf{SLDS Dataset Release:} We introduce and publicly release the \gls{SLDS} dataset, a large-scale, cross-lingual legal resource. It comprises 20K rulings from the \gls{SFSC} in German, French, or Italian, each accompanied by summaries in all three languages, resulting in 60K data rows. \gls{SLDS} is openly available to support and encourage multilingual legal NLP research.
    \item \textbf{Comprehensive Benchmarking:} We fine-tune multiple models from the Qwen, Llama, and Phi families, including five Qwen variants, Llama 3.2 3B, and Phi-3.5-mini, and compare their performance to proprietary models (GPT-4o, Claude 3.5 Sonnet, and o3-mini) as well as the pre-trained DeepSeek R1 in a one-shot setting. Our evaluation, combining conventional summarization metrics with an LLM-as-a-Judge approach, highlights the trade-offs between fine-tuning and prompting while revealing the limitations of standard metrics in capturing the nuances of legal summarization.
\end{compactenum}

\begin{figure*}[tb]
    \centering
    \begin{subfigure}{0.45\textwidth}
        \centering
        \includegraphics[width=\textwidth]{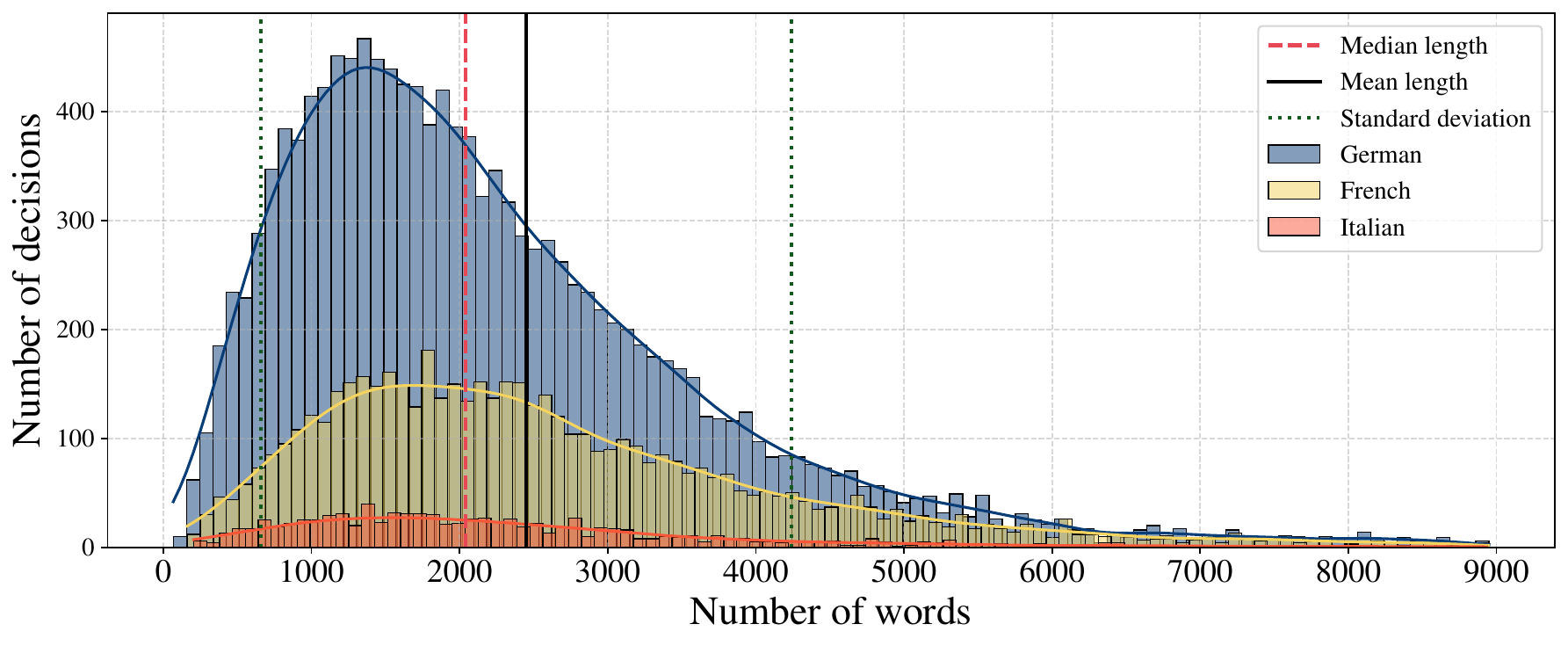}
        \caption{Decisions}
        \label{fig:lds_input_length_distribution}
    \end{subfigure}
    \begin{subfigure}{0.45\textwidth}
        \centering
        \includegraphics[width=\textwidth]{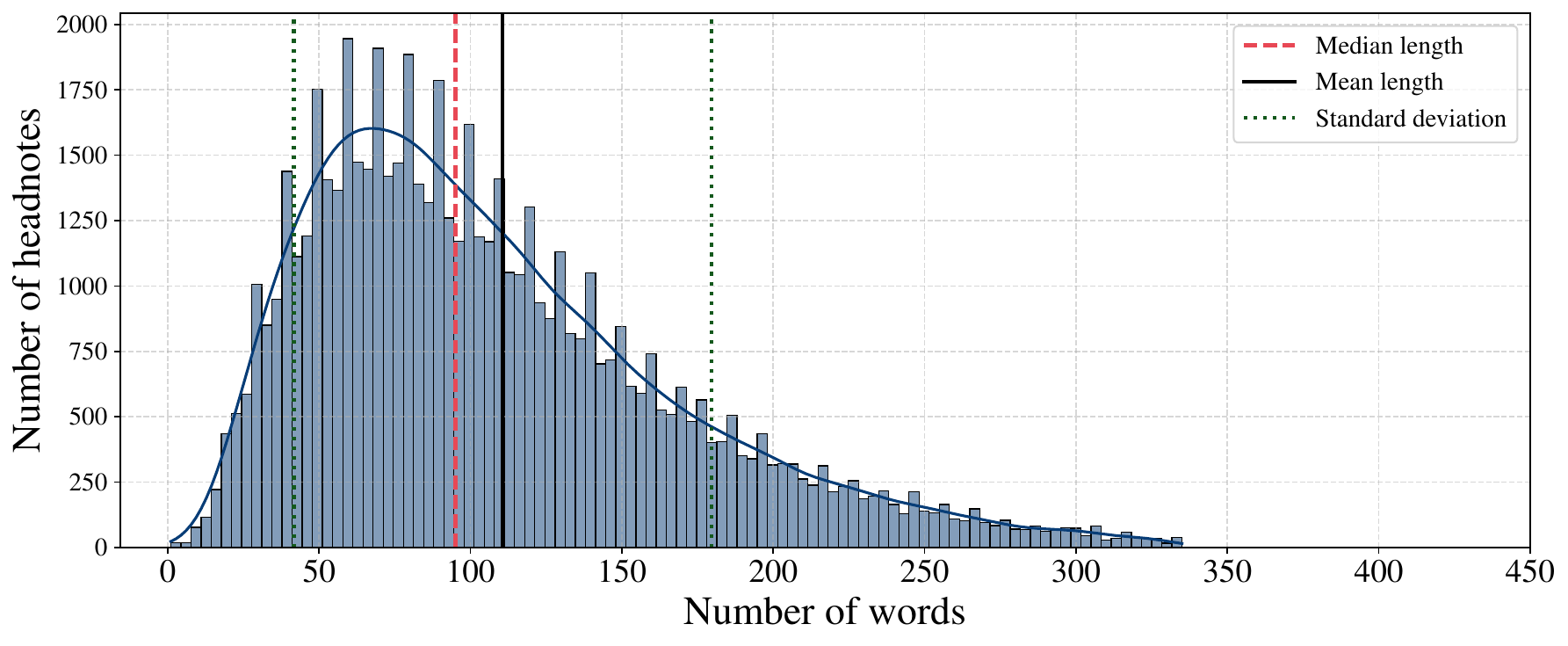}
        \caption{Headnotes}
        \label{fig:lds_summary_length_distribution}
    \end{subfigure}
    \vspace{-1ex}
    \caption{Distributions of token counts in (a) landmark decisions and (b) headnotes. To improve readability, only samples within the 99th percentile were included, as the long tail of the distribution would have otherwise skewed the visualization.}
    \vspace{-2ex}
    \label{fig:lds_length_distribution}
\end{figure*}

%% file: sections/related_work.tex
\section{Related Work}

Research on legal text summarization has increasingly shifted toward abstractive methods, supported by the emergence of dedicated datasets. Among monolingual English corpora, BillSum \citep{kornilova2019billsum} offers 22K U.S. congressional and state bills with summaries, enabling cross-domain transfer for legal summarization. Multi-LexSum \citep{shen_multi-lexsum_2022} focuses on long civil rights lawsuits and supports multi-length evaluations. \citet{bauer2023legal} extracted key passages from 430K U.S. court opinions, favoring reinforcement learning methods, although their dataset is not publicly available. RulingBR \citep{de2018rulingbr} includes 10K Brazilian Supreme Court rulings with structured summaries. LAWSUIT \citep{ragazzi2024lawsuit} contains 14K Italian verdicts with expert-authored maxims from the Constitutional Court.

Multilingual datasets include EUR-Lex-Sum \citep{aumiller_eur-lex-sum_2022}, which covers 24 EU languages and aligns 375 legal acts. Unlike court rulings, these acts follow a more structured format. In contrast, our dataset emphasizes case law within a single jurisdiction, offering over 13 times more French-to-German and more than twice as many Italian-to-German examples than EUR-Lex-Sum. MILDSum \citep{datta-etal-2023-mildsum} addresses language barriers in India by translating 3K English judgments to Hindi. A key result was that Summarize-then-Translate outperformed direct cross-lingual summarization. Unlike MILDSum, our dataset excludes English and uses headnotes, which are harder to generate than summaries due to their legal specificity and structural requirements, making the task more challenging given the dominance of English in pretraining corpora.

%% file: sections/dataset.tex
\begin{figure*}[tb]
    \centering
    \includegraphics[width=0.95\textwidth]{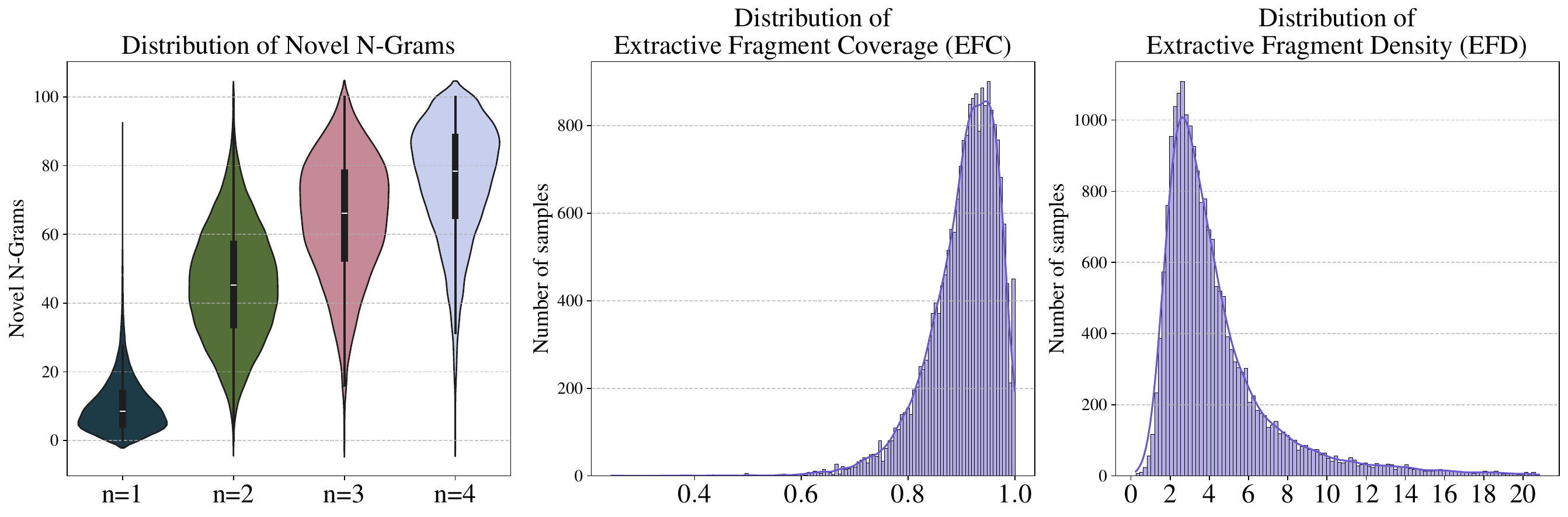}
    \vspace{-1ex}
    \caption{Distribution of Summarization Properties in SLDS. The figure illustrates N-Gram Novelty (left), \gls{EFC} (center), and \gls{EFD} (right), highlighting the dataset's balance between abstraction and extractiveness. For the sake of readability, we only show \gls{EFD} values within the 99th percentile.}
    \vspace{-2ex}
    \label{fig:summarization_metrics_violin_plots}
\end{figure*}

\begin{figure}[!htb]
    \begin{judgeExampleSmall}
        {61194}
        {Claude 3.5 Sonnet}
        {\input{examples/judge/61194/decision_translated_excerpt}}
        {\input{examples/judge/61194/original_headnote_translated_excerpt}}
        {\input{examples/judge/61194/generated_headnote_translated_excerpt}}
        {40}
        {\input{examples/judge/61194/judge_output_excerpt}}
    \end{judgeExampleSmall}
    \vspace{-3ex}
    \caption{Example of a DeepSeek V3 judgment for a headnote generated by Claude 3.5 Sonnet. The full sample can be seen in \Cref{app:examples:judge}.}
    \label{fig:judge-example-output-excerpt}
    \vspace{-3ex}
\end{figure}

\section{Data}

We introduce \gls{SLDS}, a novel dataset for cross-lingual summarization in the legal domain. It comprises over 20K landmark decisions published by the \gls{SFSC} in German, French, or Italian, each accompanied by paragraph-aligned summaries written by clerks and judges in all three languages. This dataset provides a valuable resource for studying cross-lingual summarization, a relatively underexplored area in legal NLP. Unlike datasets such as EUR-Lex-Sum, which focus on legislation, \gls{SLDS} centers on judicial decisions, making it particularly relevant for developing tools to assist legal practitioners and researchers working with court rulings.

\subsection{Data Collection}

We scraped the decisions from the official Swiss Federal Supreme Court repository, covering 70 years and five legal volumes.\footnote{Available at \url{https://www.bger.ch/}} We extracted the full decision text, either in German, French or Italian, along with the headnotes in all three languages. We also stored and inferred metadata including the year of the decision, the volume in which the decision was published, the law area of the decision which can be inferred from the volume and the year, and the url to the official published decision on the repository. To enable model training and cross-lingual evaluation, each row contains one decision-headnote pair, tripling the dataset to over 60K samples. We show the exact fields of our dataset in \Cref{app:dataset:fields}.

\subsection{General Information}
\label{sec:dataset-stats}

\paragraph{Dataset Splits}
The dataset is partitioned by publication year to prevent data leakage and maintain consistency with current summarization styles. As shown in \Cref{tab:dataset_splits}, the training set spans 1954–2021, the validation set covers 2022, and the test set includes 2023–2024, resulting in approximately 60K, 600, and 978 samples per split. For a detailed year-wise distribution, see \Cref{app:dataset:lds_year_distribution}.

\begin{table}[h]
    \centering
    \vspace{-1ex}
    \resizebox{\columnwidth}{!}{%
        \begin{tabular}{lrrrc}
            \toprule
            \textbf{Split} & \textbf{Years} & \textbf{\# Dec.} & \textbf{\# Samp.} & \textbf{Languages (\%)}        \\
            \midrule
            Train          & 1954–2021      & $\sim$20k          & $\sim$60k        & DE: 67.94, FR: 27.36, IT: 4.71 \\
            Val     & 2022           & 200                   & 600                 & DE: 68.50, FR: 27.50, IT: 4.00 \\
            Test           & 2023–2024      & 326                   & 978                 & DE: 63.50, FR: 32.82, IT: 3.68 \\
            \bottomrule
        \end{tabular}
    }
    \caption{Dataset splits by publication years and language distribution of decisions (Dec).}
    \label{tab:dataset_splits}
    \vspace{-3ex}
\end{table}

\paragraph{Text Length}
\autoref{fig:lds_length_distribution} shows the number of tokens for both decisions and headnotes up to the 99th percentile. Decisions range from 102 to 44.3K tokens. The median decision length is 2971 tokens, and the mean decision length is 3585 tokens with a standard deviation of 2629 tokens.

\subsection{Summarization-related Properties}

To analyze the summarization tendencies in SLDS, we examine Compression Ratio (CR), Extractive Fragment Coverage (EFC), Extractive Fragment Density (EFD) \citep{grusky-etal-2018-newsroom}, and N-Gram Novelty \citep{narayan-etal-2018-dont}. Given the dataset's multilingual nature, we compare these properties to EUR-Lex-Sum \cite{aumiller_eur-lex-sum_2022} and MILDSum \cite{datta-etal-2023-mildsum}, but only for monolingual samples. We also report \gls{CI} and Formulaicness \citep{ragazzi2024lawsuit}. \autoref{fig:summarization_metrics_violin_plots} visualizes key trends across the entire dataset.

\paragraph{Compression Ratio}

We compute \gls{CR} as the ratio of decision to headnote token counts, using language-specific \texttt{spaCy} tokenizers via the \texttt{spacy.blank} interface. The observed mean \gls{CR} of 26.39 is notably higher than in EUR-Lex-Sum and MILDSum, reflecting the extreme conciseness of Swiss judicial headnotes. These headnotes highlight key legal principles that justify a decision's landmark status. Higher \glspl{CR} in the validation and test splits suggest a trend toward even shorter headnotes over time.

\paragraph{Extractive Fragments}

We compute \gls{EFC} and \gls{EFD} using spaCy \citep{spacy} with \verb|core_news_sm| models on monolingual samples. While \gls{EFC} values match MILDSum, this may result from longer input texts and high CRs, which increase unigram overlap. The mean \gls{EFD} of 4.63, however, is significantly lower than MILDSum's 24.42, indicating a more abstractive summarization style. Slightly higher \gls{EFC} and \gls{EFD} values in the validation/test sets align with their increased \glspl{CR}.

\paragraph{N-Gram Novelty}

We measure abstractivity by the proportion of novel n-grams in headnotes versus source decisions \citep{narayan-etal-2018-dont}. Novelty ranges from 0 (fully extractive) to 100 (fully abstractive). On average, about 90\% of headnote unigrams appear in the decision, and only ~5\% are novel in the test set. Novelty increases with longer n-grams, suggesting that headnotes often reuse the same words but in new combinations. Around 30\% of quadgrams are copied verbatim, highlighting the dataset's mix of extractive and abstractive styles. Compared to the unigram novelty of around 40\% and bigram novelty of 64-67\% reported for the German, French, and Italian subsets of EUR-Lex-Sum, the headnotes in \gls{SLDS} adhere more closely to the original wording of the source decisions. More detailed statistics are in \Cref{app:summarization-properties}.

\paragraph{Coverage Increment and Formulaicness}

\begin{figure*}[tb]
    \centering
    \includegraphics[width=0.85\textwidth]{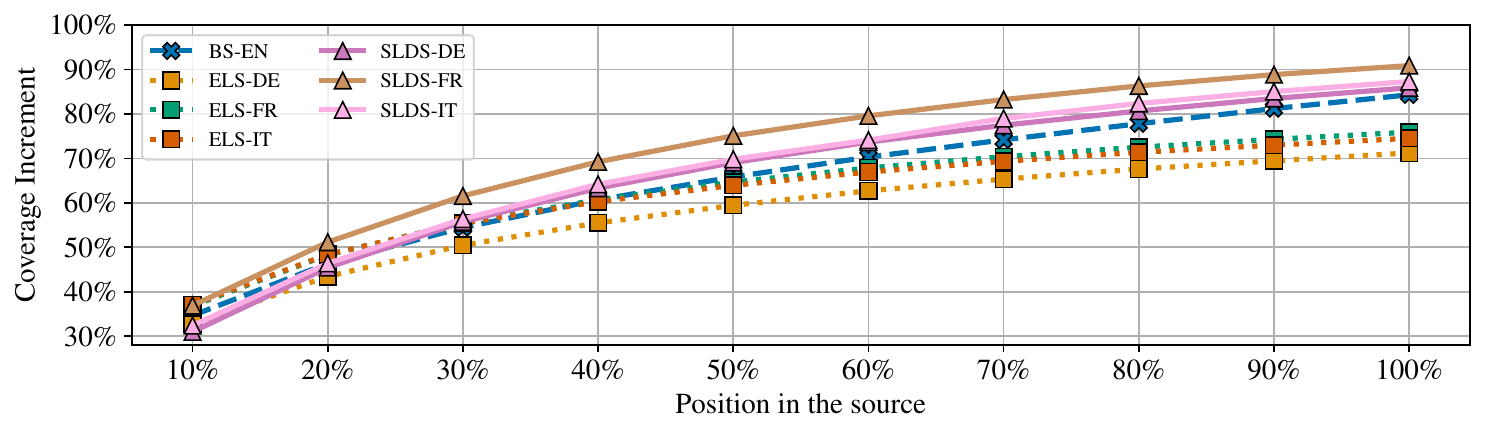}
    \vspace{-2ex}
    \caption{Percentage of unique unigrams in the headnote that also appear in the decision text, reported for the monolingual German, French, and Italian subsets in SLDS and EUR-Lex-Sum (ELS), and for the entire English dataset in BillSum (BS-EN).}
    \vspace{-1ex}
    \label{fig:coverage_increment}
\end{figure*}

Following \citet{ragazzi2024lawsuit}, we compute \gls{CI} and Formulaicness on monolingual samples. To obtain CI, we divide each decision into ten equal-length segments and compute the proportion of headnote unigrams that also appear in each segment. \Cref{fig:coverage_increment} shows that SLDS exhibits \gls{CI} values similar to BillSum, especially in German and Italian \gls{SLDS} samples, while EUR-Lex-Sum displays slightly lower values, in line with its higher abstractivity.

\begin{figure}[tb]
    \centering
    \includegraphics[width=\linewidth]{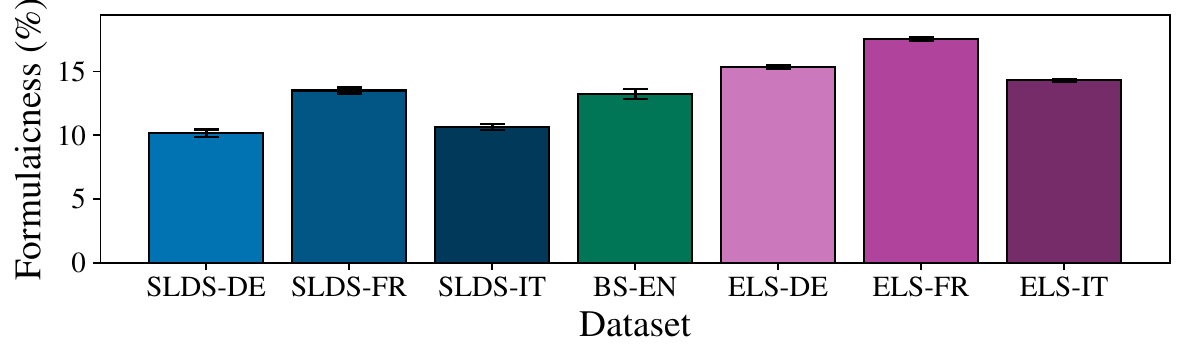}
    \vspace{-4ex}
    \caption{Average headnote formulaicness reported for the monolingual German, French, and Italian subsets in SLDS and EUR-Lex-Sum (ELS), and for the entire English dataset in the case of BillSum (BS-EN).}
    \label{fig:formulaicness}
    \vspace{-3ex}
\end{figure}

Formulaicness is computed by averaging ROUGE-L F1 scores between headnotes across random subsets. \Cref{fig:formulaicness} shows that SLDS-DE and SLDS-IT have the lowest scores, indicating greater variability in phrasing. The French subset is similar to BillSum in this regard, while EUR-Lex-Sum exhibits the highest Formulaicness despite its higher N-Gram Novelty. This finding supports the hypothesis that \gls{SLDS} headnotes, although largely composed of words found in the original decisions, frequently rearrange these words in novel ways. As a result, they strike a distinctive balance between extractiveness and abstractiveness.

\subsection{Licensing}
We release the dataset under the CC-BY-4.0 license, which complies with the \gls{SFSC} licensing.\footnote{For more information, see \url{https://www.bger.ch/files/live/sites/bger/files/pdf/de/urteilsveroeffentlichung_d.pdf}}

\subsection{Ethical Considerations}
Due to the sensitive nature of court cases and their corresponding rulings, the \gls{SFSC} anonymizes personal or sensitive information according to their guidelines before publishing them online.\footnote{Anonymization guidelines at \url{https://www.bger.ch/home/juridiction/anonymisierungsregeln.html}}

%% file: examples/judge/61194/decision_translated_excerpt.tex
150 III 223 Facts from page 225 \\
A. A. was pursued for a claim of CHF 200 and a reminder fee of CHF 35 (debt collection no. w by the Zug Debt Collection Office) (…) The Federal Supreme Court has already dealt with the costs for a pickup invitation in the past. (...) Therefore, the insufficiently substantiated complaint is not to be admitted (cf. unpublished E. 1).

%% file: examples/judge/61194/original_headnote_translated_excerpt.tex
Art. 1, Art. 2, Art. 9 para. 1 let. a, Art. 10bis, (...); Costs of payment orders, seizure notices, and loss certificates. General principles on fees and compensations according to the GebV SchKG (consid. 3.1).
Costs for the delivery of payment orders (consid. 3.2.1);
(...)

%% file: examples/judge/61194/generated_headnote_translated_excerpt.tex
**Art. 9, 13, 15, 20, and 10bis GebV SchKG; Fees and compensations in debt enforcement proceedings.**
In addition to the fee under Art. 16 para. 1 GebV SchKG, expenses for postal charges (Art. 13 para. 1 GebV SchKG) may be charged for the delivery of a payment order.
(...)

%% file: examples/judge/61194/judge_output_excerpt.tex
ACCURACY\_FAITHFULNESS: \\
Analysis: The Model-Generated Headnote captures the general legal principles and main points of the Official Headnote but lacks some specific details and precise references. For instance, it omits references to Articles 1, 2,  (...) \\
ACCURACY\_FAITHFULNESS\_SCORE: 2 \medskip

(Other judge categories skipped for brevity)

%% file: sections/setup.tex
\begin{table*}[t]
    \centering
    \tiny
    \vspace{-1ex}
    \resizebox{\textwidth}{!}{
        \begin{tabular}{lrrrrrrrr}
            \toprule
            \textbf{Model}    & \textbf{Setting} & \textbf{BERTScore} $\uparrow$     & \textbf{BLEU} $\uparrow$          & \textbf{ROUGE-1} $\uparrow$       & \textbf{ROUGE-2} $\uparrow$       & \textbf{ROUGE-L} $\uparrow$       & \textbf{JUDGE} $\uparrow$         \\
            \midrule
            Phi-3.5-mini      & fine-tuned       & 11.24 ± 3.82                      & 34.84 ± 0.41                      & 31.20 ± 2.08                      & 14.11 ± 1.27                      & 20.96 ± 1.35                      & 15.25 ± 2.32                      \\
            Llama 3.2 3B      & fine-tuned       & 15.20 ± 4.40                      & 21.89 ± 0.42                      & 31.89 ± 2.34                      & 14.87 ± 1.61                      & 22.49 ± 1.60                      & 18.47 ± 2.99                      \\
            Qwen2.5 0.5B      & fine-tuned       & -1.37 ± 3.85                      & 32.20 ± 0.35                      & 23.87 ± 1.68                      & 9.46 ± 0.94                       & 17.37 ± 1.09                      & 5.80 ± 1.26                       \\
            Qwen2.5 1.5B      & fine-tuned       & 19.81 ± 2.72                      & 36.79 ± 0.34                      & 33.03 ± 1.73                      & 14.14 ± 1.08                      & 22.67 ± 1.13                      & 15.92 ± 2.27                      \\
            Qwen2.5 3B        & fine-tuned       & 23.23 ± 2.80                      & 38.42 ± 0.34                      & 35.18 ± 1.79                      & 15.66 ± 1.23                      & 24.10 ± 1.17                      & 20.31 ± 2.66                      \\
            Qwen2.5 7B        & fine-tuned       & 29.59 ± 1.97                      & 41.40 ± 0.34                      & 39.24 ± 1.59                      & 18.26 ± 1.25                      & 26.44 ± 1.15                      & 28.37 ± 3.07                      \\
            Qwen2.5 14B       & fine-tuned       & \textbf{\underline{32.48 ± 1.98}} & \textbf{\underline{41.80 ± 0.37}} & \underline{40.04 ± 1.74}          & \textbf{\underline{19.99 ± 1.41}} & \textbf{\underline{28.00 ± 1.28}} & \underline{31.38 ± 3.19}          \\
            \midrule
            GPT-4o            & one-shot         & \underline{30.44 ± 1.74}          & \underline{31.89 ± 0.25}          & \textbf{\underline{42.12 ± 1.79}} & 18.92 ± 1.22                      & 25.92 ± 1.05                      & 39.70 ± 2.66                      \\
            Claude 3.5 Sonnet & one-shot         & 5.53 ± 2.00                    & 21.88 ± 0.25                      & 41.86 ± 1.64                      & \underline{19.23 ± 1.19}          & \underline{27.67 ± 1.20}          & 41.25 ± 2.90                      \\
            DeepSeek-R1       & one-shot         & 20.28 ± 1.45                      & 22.37 ± 0.18                      & 38.30 ± 1.82                      & 15.97 ± 0.85                      & 21.03 ± 0.84                      & \textbf{\underline{42.28 ± 2.21}} \\
            o3-mini           & one-shot         & 14.18 ± 1.31                      & 20.55 ± 0.17                      & 34.77 ± 1.43                      & 11.92 ± 0.69                      & 18.21 ± 0.67                      & 34.82 ± 2.41                      \\
            \bottomrule
        \end{tabular}
    }
    \vspace{-1ex}
    \caption{Baselines on the \gls{SLDS} test set, macro-averaged over the nine decision and headnote language combinations. Standard errors are estimated with bootstrapping as implemented in lighteval \citep{lighteval}. For BERTScore we report the F1 score. The ROUGE scores are multiplied by 100 for consistency. \textbf{Bold}: best overall; \underline{underlined}: best within setup.}
    \label{tab:results-overall}
    \vspace{-4ex}
\end{table*}

\section{Experimental Setup}

We evaluate four frontier \glspl{LLM} (GPT-4o, Claude~3.5~Sonnet, DeepSeek~R1, o3-mini) in a one-shot setting and fine-tune three \glspl{SLM} (Llama~3.2~3B, Qwen2.5~3B, and Phi-3.5-mini) on the SLDS training split. To assess the effect of model size, we fine-tune additional Qwen2.5 variants (0.5B~-~14B) and evaluate them in a zero-shot setting.\footnote{Models are available on Hugging Face: \url{https://huggingface.co/ipst}} Appendix \ref{app:experiment-details} details the model versions, decoding parameters, and one-shot prompting. Fine-tuning hyperparameters are listed in \Cref{app:hyperparameters-fine-tuning}.

\subsection{Traditional Metrics}

We evaluate models on the \gls{SLDS} test set using \texttt{lighteval} \citep{lighteval}, reporting BERTScore \citep{zhang_bertscore_2020}, BLEU \citep{BLEU}, and ROUGE \citep{lin_rouge_2004}. Since each metric has known limitations \citep{zhang_bertscore_2020}, we report all three to capture complementary aspects of summarization performance.\footnote{The evaluation script is available at \url{https://github.com/rolshoven/slds-eval}} For more details on the hyperparameters we used in certain metrics, refer to \Cref{app:metric-hyperparams}.

\subsection{LLM-as-a-Judge}

We further adopt the LLM-as-a-Judge framework \citep{zheng2023judging}, employing DeepSeek V3 \citep{liu2024deepseek} as the judge model due to its multilingual capabilities, low cost, and the fact that it was not among the evaluated models, avoiding bias toward self-generated outputs \citep{panickssery2024llm}. To enable a fair comparison with human evaluation, neither the LLM nor the human judges were shown the full decision text. Given the high quality of the gold headnotes, this setup provides a meaningful and token-efficient evaluation.

\paragraph{Evaluation Protocol} The judge LLM evaluates generated headnotes against gold headnotes across five dimensions: (1) Accuracy \& Faithfulness, (2) Completeness \& Relevance, (3) Clarity \& Coherence, (4) Articles (whether legal articles are correctly and completely referenced), and (5) Considerations (whether the correct legal considerations are identified and preserved). It provides a short analysis and assigns a score from 1 (major flaws) to 3 (close match) per category. Prompts and an example output are shown in Appendix \ref{app:prompts:judge} and \ref{app:examples:judge}. An example is shown in \cref{fig:judge-example-output-excerpt}. For the full texts, refer to \Cref{app:examples:judge}. 

While we cannot directly use existing tools such as AlignScore \citep{zha-etal-2023-alignscore} or SummaC \citep{laban-etal-2022-summac} to check for factual consistency of the generated headnotes in our cross-lingual setup without significant adaptation, our \textit{Completeness \& Relevance} can serve as a proxy for factual consistency as it penalizes any deviation from the gold-standard headnote, effectively punishing factual inaccuracies and hallucinations.


\paragraph{Aggregation} To compute the final score, each rating is normalized from 1-3 to 0-2. The five normalized scores for a sample are summed (max 10) and multiplied by 10 to yield a percentage between 0 and 100. We did not apply weighting, as experts deemed all rubrics equally important. The final judge score is the average of these scaled values across all test samples.

\subsection{Human Evaluations}

To obtain a trusted qualitative estimate of model performance, we sampled 63 instances from the test set, with seven per decision-headnote language pairs across all nine subsets (such as de→de, de→fr), resulting in a total of 189 generated headnotes evaluated against 63 gold headnotes. Each sample included the original headnote and outputs from the top-performing models in three categories: fine-tuned, frontier, and reasoning models. Two co-authors, both professional lawyers fluent in the relevant languages, assessed the samples using the same protocol as the \gls{LLM} judge. We prioritized broader coverage across all language pairs in the dataset over inter-annotator agreement, especially since only one of the two experts is fluent in Italian. Expanding the evaluation further was not feasible due to the high cost of legal expertise, while using less experienced annotators would have compromised quality. Additionally, a third legal expert and co-author of this paper conducted an in-depth qualitative analysis of six selected samples, taking into account the full decision text. This setup provided valuable expert insights while balancing quality and feasibility.

%% file: sections/results.tex
\begin{figure*}[tb]
    \centering
    \includegraphics[width=0.9\textwidth]{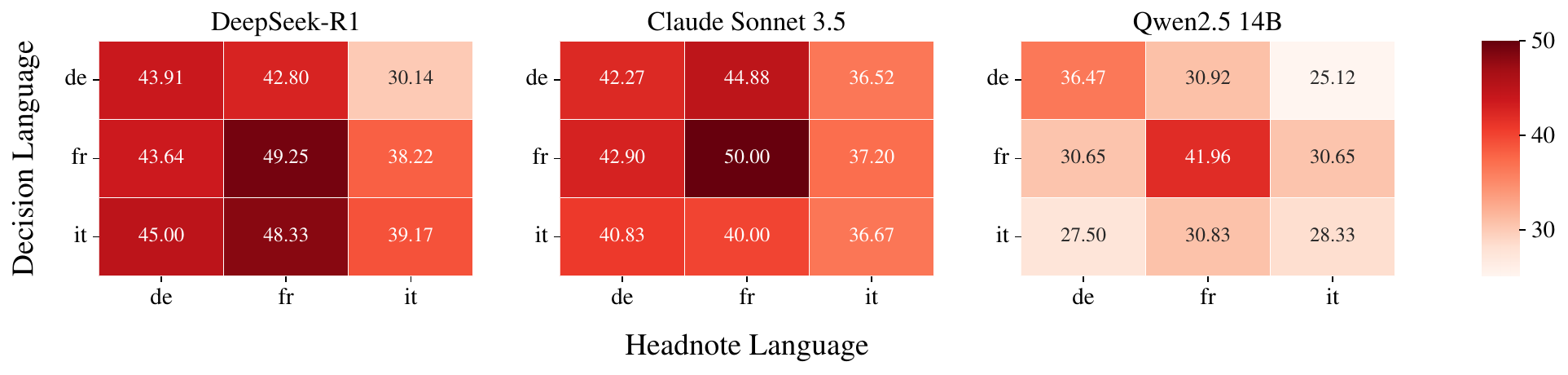}
    \vspace{-2ex}
    \caption{JUDGE scores for different cross-lingual language subsets and different models. Darker colors indicate better scores.}
    \label{fig:cross-lingual-heatmap}
    \vspace{-3ex}
\end{figure*}

\begin{figure*}[htb]
    \centering
    \includegraphics[width=0.9\textwidth]{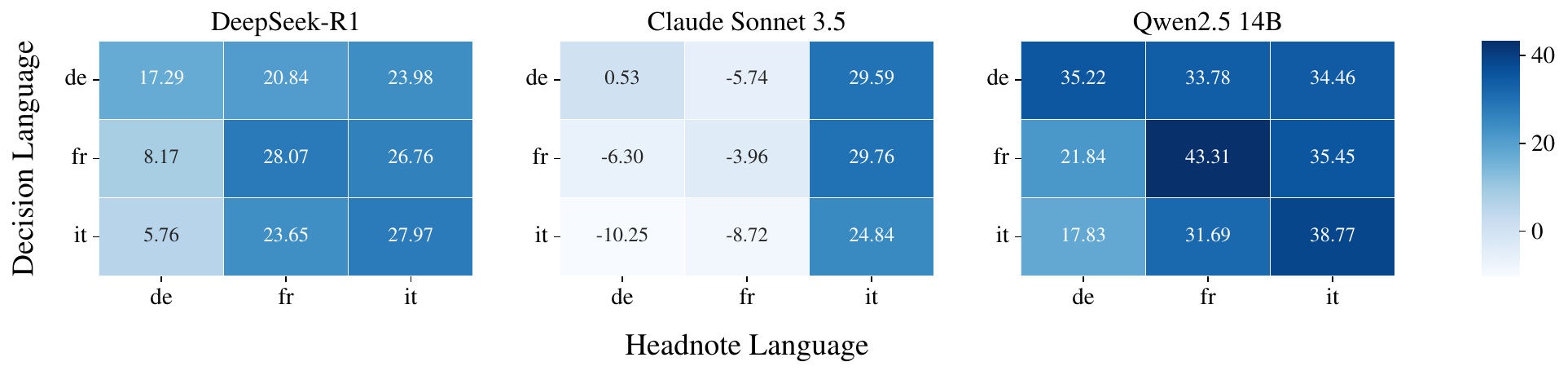}
    \vspace{-2ex}
    \caption{BERTScore for different cross-lingual language subsets and different models. Darker colors indicate better scores.}
    \label{fig:cross-lingual-heatmap-bertscore}
    \vspace{-2ex}
\end{figure*}

\section{Results}

\subsection{Overall Results}

We present the results of our evaluations on the \gls{SLDS} test set in \Cref{tab:results-overall}. We macro-averaged over the scores in each of the nine language subsets of decision and headnote language pairs to promote fairness and robustness across languages. Below, we highlight several interesting observations. For Claude 3.5 Sonnet, post-processing of the BERTScore metric was necessary. We discovered that four samples with empty outputs were assigned extremely negative scores due to headnotes that were not generated, which skewed the aggregate result. We corrected this by neutralizing the scores for these samples. A detailed explanation of this procedure is available in \Cref{app:claude-negative-bertscores}.

\paragraph{Fine-tuned models perform well on automated metrics, but lag in legal precision.}
Although smaller \glspl{SLM} achieve lower JUDGE scores than their larger counterparts, our results show that the fine-tuned Qwen2.5 14B surpasses even significantly larger proprietary models on standard metrics such as BERTScore, BLEU, ROUGE-2, and ROUGE-L. ROUGE-1 scores for Qwen are also notably high. This indicates that the fine-tuned models excel in lexical similarity but still fall short in legal correctness, completeness, and structural fidelity when compared to large proprietary \glspl{LLM}. These findings highlight the limitations of traditional automated metrics and emphasize the need for more sophisticated evaluation methods based on \glspl{LLM} as judges. Nevertheless, fine-tuning on the SLDS training split leads to a substantial improvement in JUDGE scores on the test set, as illustrated in \Cref{app:off-the-shelf-vs-finetuned}.

\paragraph{Large models are more accurate.} Our results indicate that larger models are better at generating headnotes that are legally accurate, complete and faithful, as indicated by the higher judge scores. While this was expected, we hypothesize that it could be partially due to the one shot examples provided in the prompt. Although we initially considered one-shot prompting for the fine-tuned models, it did not improve performance, likely because these models had already learned the headnote format during training. Another interesting observation is that Claude~3.5~Sonnet performs second best in the judge score but has the lowest BERTScore in the one-shot setting. This shows that certain metrics can be deceptive and that relying on a single metric for evaluating summaries is usually not sufficient. In the case of Claude 3.5 Sonnet, the low BERTScores are likely attributable to the model's tendency to produce an unrequested JSON output format, which penalized the token-based comparison. Further details on this output behavior can be found in \Cref{app:claude-negative-bertscores}.

\paragraph{Reasoning Models Offer Limited Gains} Interestingly, the reasoning models do not perform significantly better. Even though DeepSeek~R1 outperforms all other models in terms of the JUDGE score, the difference to Claude 3.5 Sonnet is only one point. Moreover, o3-mini performs worse than Claude 3.5 Sonnet and only slightly outperforms our fine-tuned Qwen2.5 14B model by roughly 3.4 points. Our findings suggest that generating legal headnotes primarily requires factual accuracy, domain knowledge, and structured outputs, rather than complex logical reasoning. The task primarily demands models to faithfully extract and concisely rephrase key legal principles, ensuring that references to legal articles and considerations remain intact. Given that general-purpose models such as GPT-4o and Claude 3.5 Sonnet achieve similar or better judge scores than reasoning models, this indicates that current LLMs already possess sufficient reasoning capabilities for this summarization task.

\subsection{Cross-lingual Subsets}
We report cross-lingual results based on the decision and headnote language (\textit{subsets}), e.g., \verb|de_fr| for decisions in German with French headnotes. Key findings are summarized below with full details in Appendix \Cref{tab:results-cross-lingual}. To facilitate the analysis of JUDGE and BERTScores in the cross-lingual settings, we provide heatmaps of selected models in \Cref{fig:cross-lingual-heatmap} and \Cref{fig:cross-lingual-heatmap-bertscore}.

\paragraph{Qwen2.5 14B struggles with cross-lingual consistency.} While Qwen2.5 14B performs well in monolingual French (fr→fr), its scores drop significantly when the headnote language differs from the decision language, particularly for German and Italian sources. This suggests \emph{weaker cross-lingual robustness} despite strong monolingual performance.

\paragraph{French headnotes often score highest.} French headnotes tend to achieve higher JUDGE scores, particularly in the monolingual fr→fr setting. This trend also appears frequently, though not universally, in cross-lingual cases such as de→fr with Claude~3.5~Sonnet, it→fr with DeepSeek~R1, and it→fr with Qwen2.5~14B. In cases where French is not the top-performing target language, the score differences are usually small. This may suggest either higher model proficiency in generating French legal text or that French headnotes are more systematically structured and easier to reproduce.

\paragraph{Limitations of general-purpose metrics.}
The heatmaps in \Cref{fig:cross-lingual-heatmap,fig:cross-lingual-heatmap-bertscore} reveal substantial inconsistencies between the two metrics. Some model outputs from Claude 3.5 Sonnet receive low BERTScores while achieving high JUDGE scores, indicating strong performance in legal correctness, completeness, and clarity. These observations show the limitations of general-purpose similarity metrics and emphasize the need for \emph{domain-specific evaluation methods} in legal text generation.

\begin{figure*}[tb]
    \centering
    \begin{subfigure}{0.45\textwidth}
        \centering
        \includegraphics[width=\textwidth]{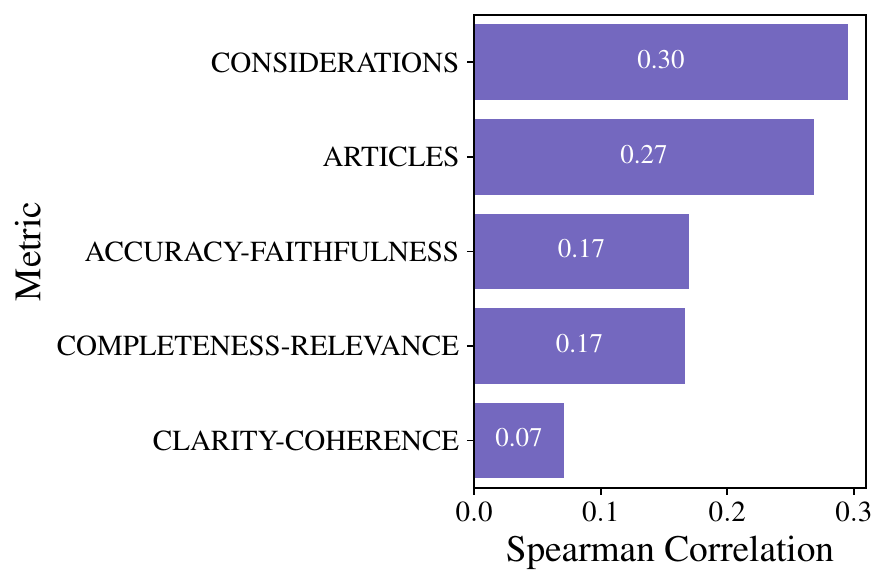}
        \caption{Human Rubric to LLM Judge Rubric}
        \label{fig:human_evaluation_correlations:categories}
    \end{subfigure}
    \begin{subfigure}{0.45\textwidth}
        \centering
        \includegraphics[width=\textwidth]{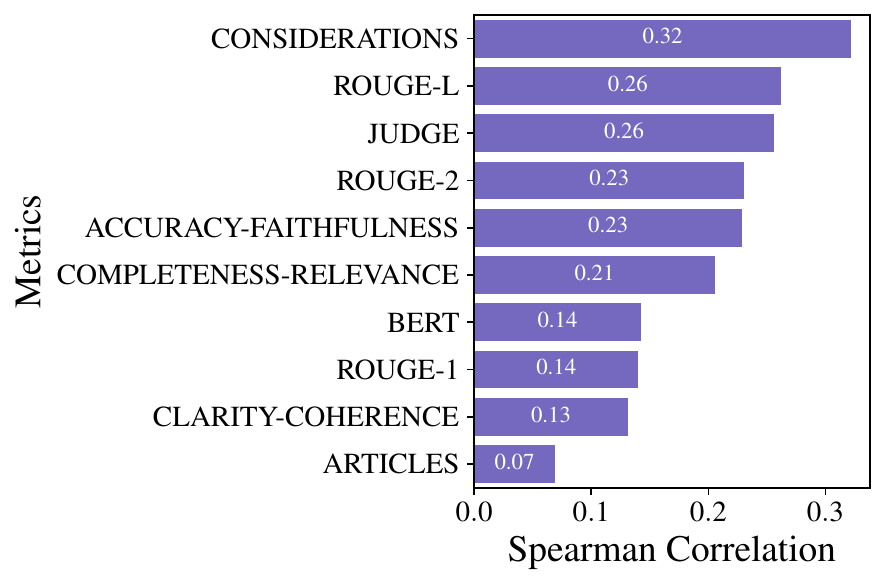}
        \caption{Overall Human Score to Rubrics}
        \label{fig:human_evaluation_correlations:metrics}
    \end{subfigure}
    \vspace{-1ex}
    \caption{
    While \gls{LLM} scores vary across categories, the overall JUDGE score remains highly correlated with human judgment. Notably, the considerations score, shows the strongest correlation with aggregated human scores.}
    \vspace{-3ex}
    \label{fig:human_evaluation_correlations}
\end{figure*}

%% file: sections/expert_evaluation.tex
\section{Human Expert Evaluation}

We perform two human expert evaluations. The first is based on the same evaluation process the \gls{LLM} judge also follows. Two lawyers assess three generated headnotes across 63 samples. This evaluation only considers the generated and the original headnote without taking into account the actual text of the landmark decision, assuming the gold headnote is the ideal headnote and that any deviation should be penalized. We refer to this evaluation as \textit{Human-as-a-Judge}. In the second evaluation which we will refer to as \textit{Contextualized Human Analysis}, another lawyer looked at six of those 63 samples and performed an in-depth analysis which involved studying the decision text as well.

\subsection{Human-as-a-Judge}

With 63 decisions and headnotes generated by three models, we obtained 189 annotated samples. Appendix \Cref{fig:judge_and_human_score_distributions} illustrates score distributions assigned by both the \gls{LLM} and the lawyers. The latter tend to give slightly higher scores than DeepSeek-V3, with a mean difference of 11.64, indicating that the \gls{LLM} judge is stricter in its assessments.

\paragraph{Evaluation Metrics} \Cref{fig:human_evaluation_correlations} presents two correlation analyses assessing our legal headnote evaluation. \Cref{fig:human_evaluation_correlations:categories} shows Spearman correlations between DeepSeek-V3’s category-specific scores and human expert ratings across five dimensions. \Cref{fig:human_evaluation_correlations:metrics} compares traditional metrics (ROUGE, BERTScore) and \gls{LLM}-based judgments with aggregated human scores. We present our findings in the following paragraphs. 

\paragraph{Correlation Analysis} \Cref{fig:human_evaluation_correlations} reveals important patterns in how automated evaluation approaches align with human judgment. Examining the category-wise correlations in \Cref{fig:human_evaluation_correlations:categories}, we find that objective elements of legal analysis show the strongest agreement between human and \gls{LLM} evaluators. The \textit{Considerations} and \textit{Articles} categories demonstrate the highest correlations (0.30 and 0.27 respectively), suggesting that \glspl{LLM} are most reliable when evaluating concrete, verifiable aspects of legal headnotes. However, the markedly lower correlation in \textit{Clarity \& Coherence} (0.07) highlights a crucial limitation: automated systems struggle to assess the more nuanced, subjective aspects of legal writing that human experts evaluate with ease.

\paragraph{Metric Comparison} The analysis of different evaluation metrics in \Cref{fig:human_evaluation_correlations:metrics} reveals the complementary strengths of traditional and \gls{LLM}-based evaluation approaches. While ROUGE-L and the overall JUDGE score show moderate correlation with human assessment (both at 0.26), the distribution of correlations across metrics suggests that no single automated measure fully captures the complexity of human evaluation. Traditional metrics like BERTScore and ROUGE variants (ranging from 0.14 to 0.26) perform comparably to \gls{LLM}-based assessments, indicating that the challenges in automated evaluation persist even with advanced language models. This finding underscores the importance of combining multiple evaluation approaches when assessing legal document generation, as different metrics capture distinct aspects of document quality that align with human judgment.

\subsection{Contextualized Human Analysis}

In addition to quantitative evaluation metrics, we conducted a qualitative assessment of model-generated headnotes with a lawyer. The expert reviewed six Swiss landmark decisions along with their original headnotes and the outputs generated by Claude 3.5 Sonnet, DeepSeek R1, and our fine-tuned Qwen2.5 14B model. While all models successfully captured the general themes of the decisions, we observed significant variations in terms of reference accuracy, legal precision, and headnote appropriateness.
The expert found that DeepSeek R1 produced closely aligned headnotes to the original ones in terms of coverage and completeness, but often included excessive detail, making them more akin to case summaries than concise headnotes. Claude 3.5 Sonnet demonstrated strengths in readability and in capturing the core judgment but introduced occasional legal misinterpretations, including statements that contradicted or over-simplified aspects of the decision. Finetuned Qwen2.5~14B showed notable improvements in referencing relevant legal provisions, including the European Convention on Human Rights (ECHR), which was not cited in the original headnote but was deemed relevant. However, it also introduced incorrect legal references in some cases and sometimes inferred conclusions absent from the decision text. Additionally, all models exhibited inconsistencies in how they structured information, affecting their suitability for legal practitioners. We show an additional analysis in \Cref{app:contextualized-evaluation}.

%% file: sections/conclusions.tex
\section{Conclusions and Future Work}

We introduce \gls{SLDS}, a large-scale cross-lingual resource for judicial summarization. We benchmark fine-tuned and proprietary models, revealing a trade-off between lexical similarity and legal accuracy. While fine-tuned models perform well on traditional summarization metrics, they struggle with legal correctness, as shown by our LLM-as-a-Judge evaluation. Proprietary models demonstrated higher legal faithfulness and structured output. Notably, reasoning models did not significantly outperform general-purpose LLMs, suggesting that headnote generation requires domain-specific precision rather than complex reasoning.

%% file: sections/limitations.tex
\section*{Limitations}

Our LLM-as-a-Judge evaluation showed only a moderate correlation with human judgments, suggesting that more sophisticated prompting strategies could improve alignment in future work. Additionally, we lack Inter-Annotator Agreement, introducing potential subjectivity due to resource constraints, the high cost of legal annotations, and language barriers.

While we experimented with fine-tuned small and mid-sized models, we did not explore fine-tuning larger-scale models that benefit from scaling laws. It remains an open question whether such models could close the gap with proprietary systems while maintaining efficiency. Future research should investigate the impact of scaling laws on legal coherence and factual accuracy, as well as refine prompting techniques to enhance both headnote generation and LLM-as-a-Judge evaluation. We hope that SLDS will foster progress in multilingual legal NLP and the development of more reliable judicial summarization systems.

%% file: sections/acknowledgements.tex
\section*{Acknowledgments}

We thank the anonymous reviewers for their thoughtful comments, which improved this paper. We also acknowledge the use of UBELIX, the HPC cluster at the University of Bern, for providing computational resources for our experiments.\footnote{\url{https://www.id.unibe.ch/hpc}}

%% file: sections/appendix.tex
\section{Potential Risks}

We believe the release of \gls{SLDS} poses minimal risk. On the contrary, we expect our dataset to foster further research and encourage the development of assistive technologies that can make the work of lawyers, judges, and clerks more efficient. However, it is crucial not to rely on these summaries blindly. We recommend using such systems as tools to enhance efficiency, rather than as substitutes for human oversight. Users must ensure that the generated summaries accurately reflect the decisions and do not introduce any misleading content, since lawyers will rely on these summaries to find relevant cases faster.

\section{Use of AI Assistants}

We used ChatGPT and Gemini 2.5 Flash to improve the content of this article. It was used to rephrase certain passages, as well as to condense them to make the text less redundant and easier to understand. We carefully checked that the generated paraphrases corresponded to our own ideas and that no errors were introduced during this process.

\section{Additional Details on Dataset}

\subsection*{Dataset Creation Pipeline}

We developed a two-stage pipeline to collect and preprocess the dataset of decisions from the \gls{SFSC}. The source documents are available via the official online archive, which publishes court decisions along with headnotes in the three official languages of Switzerland (German, French, and Italian).

\paragraph{Scraping}
We implemented an asynchronous scraping script to systematically retrieve all decisions published between 1954 and 2024 across five official volumes (I–V). For each entry, the script first accesses an index page for a given year and volume, then follows hyperlinks to individual decision pages. From each decision page, we extract the decision ID, metadata, the full German headnote, and the corresponding headnotes in French and Italian via linked language-specific pages. The full decision text is extracted from the website source while removing page breaks other irrelevant elements. To ensure robustness, the script uses exponential backoff to retry failed HTTP or timeout requests. Existing data are cached to allow resumable scraping.

\paragraph{Postprocessing}
Once all raw data is collected, a postprocessing script performs several operations: (1) assignment of each decision to a \emph{law area} (e.g., civil law, criminal law) based on the year and volume, following the official classification rules \citep{merkblatt2019} and historical documentation,\footnote{Summarized at \url{https://de.wikipedia.org/w/index.php?title=Entscheidungen_des_Schweizerischen_Bundesgerichts&oldid=253293997\#Gliederung}.} (2) automatic detection of the \emph{language of the decision} using the \texttt{langdetect} library, and (3) transformation of the dataset into a long format by \emph{melting} the multilingual headnote columns into a single column with an associated language label.

The mapping from volumes to law areas is historically defined and has changed over time. Up to 1994 (volume 120), the structure included separate sub-volumes \textbf{Ia} and \textbf{Ib}, with the following assignments:

\begin{itemize}
    \item \textbf{Ia} -- Constitutional law
    \item \textbf{Ib} -- Administrative law and international public law
    \item \textbf{II} -- Civil law
    \item \textbf{III} -- Debt enforcement and bankruptcy law
    \item \textbf{IV} -- Criminal law and criminal procedure
    \item \textbf{V} -- Social security law (successor of the EVGE series, 1926--1969)
\end{itemize}

Since 1995 (volume 121), the structure has been simplified to five volumes:

\begin{itemize}
    \item \textbf{I} -- Constitutional law
    \item \textbf{II} -- Administrative law and international public law
    \item \textbf{III} -- Civil law and debt enforcement/bankruptcy
    \item \textbf{IV} -- Criminal law and criminal procedure
    \item \textbf{V} -- Social security law
\end{itemize}

\noindent These assignments were implemented programmatically using a mapping table informed by both the court’s own documentation and secondary academic references.

The dataset is split into \emph{training} (decisions from 1954 to 2021), \emph{validation} (2022), and \emph{test} (2023–2024) sets. Each entry receives a unique \texttt{sample\_id}. A predefined set of one-shot examples is additionally marked for each language pair based on the smallest sequence length in the validation set. The final dataset is then pushed to the Hugging Face Hub, including separate configurations for each decision-headnote language pair (e.g., \texttt{de\_fr}, \texttt{it\_it}).

\subsection{Fields}
\label[appendix]{app:dataset:fields}

The dataset includes the following fields:\vspace{.5ex}
\begin{compactitem}
    \setlength\itemsep{0.1ex}
    \item \verb|sample_id|: Unique identifier for a sample.
    \item \verb|decision_id|: Identifier for a specific decision. Since each decision has headnotes in three languages, this ID appears three times in the dataset.
    \item \verb|decision|: Full text of the landmark decision in either German, French or Italian.
    \item \verb|decision_language|: ISO language code of the decision (one of \verb|de|, \verb|fr|, \verb|it|).
    \item \verb|headnote|: Text of the headnote/summary, comprising:
    i) Key legal citations, including laws and prior cases,
    ii) Thematic keywords from a legal thesaurus, and
    iii) A free-form summary of key considerations.
    \item \verb|headnote_language|: ISO language code of the headnote (one of \verb|de|, \verb|fr|, \verb|it|).
    \item \verb|law_area|: Legal domain of the decision.
    \item \verb|year|: Year the decision was issued.
    \item \verb|volume|: Publication volume of the decision.
    \item \verb|url|: Link to the official decision on the \gls{SFSC} website.
\end{compactitem}
\vspace{-1ex}

\subsection{Number of landmark decisions by Year}
\label[appendix]{app:dataset:lds_year_distribution}

\begin{figure}[htb]
    \centering
    \includegraphics[width=1\linewidth]{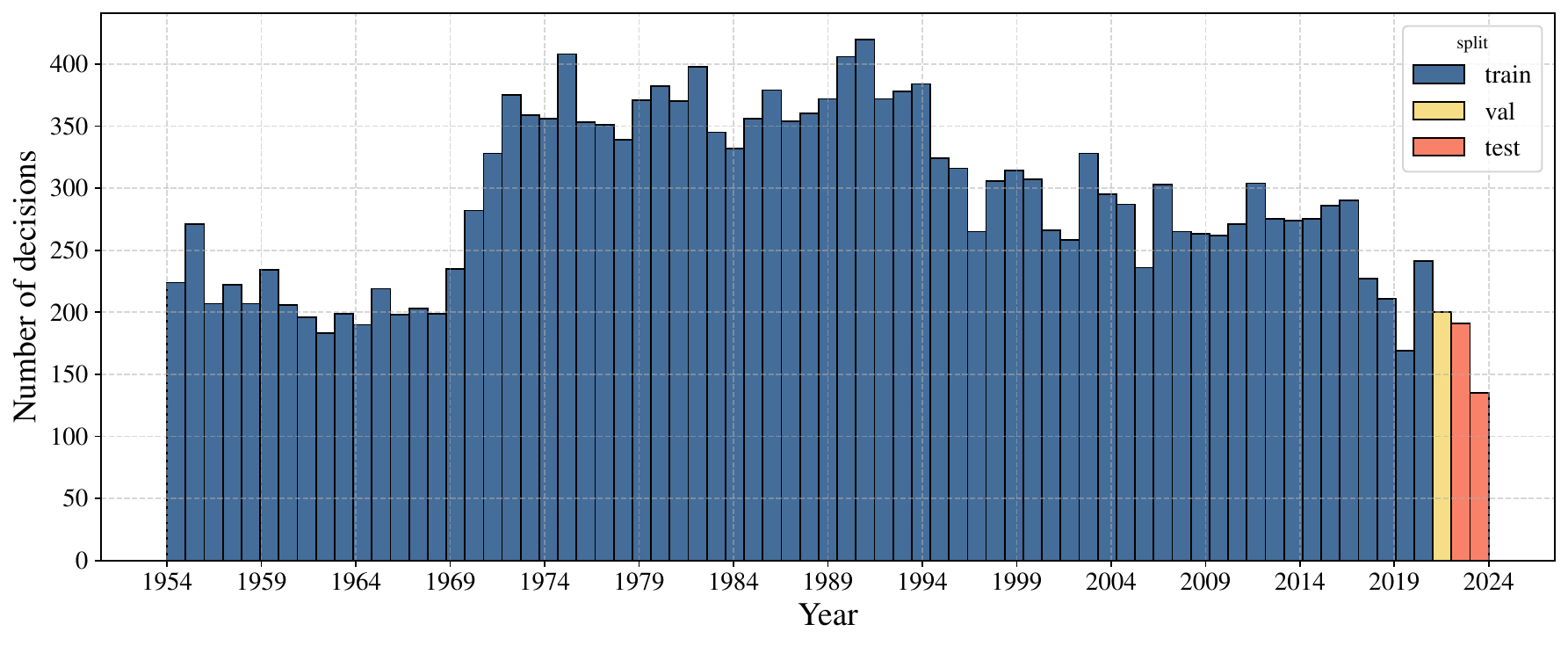}
    \caption{Number of landmark decisions published per year.}
    \label{fig:lds_year_distribution}
\end{figure}

In \Cref{fig:lds_year_distribution}, we provide a distribution of \glspl{LD} over the years.

\subsection{Properties related to Summarization}
\label[appendix]{app:summarization-properties}

We provide detailed statistics about summarization-related properties across different dataset splits in \Cref{tab:summarization_properties}.

\begin{table*}[h]
    \centering
    \small
    \begin{tabular}{llcccccc}
        \toprule
        \textbf{Metric} & \textbf{Subset} & \textbf{Mean} & \textbf{Std} & \textbf{Min} & \textbf{Median} & \textbf{Max} \\ \midrule
        CR              & Overall         & 26.39         & 30.09        & 1.89         & 21.42           & 3710.5       \\
                        & Train           & 26.21         & 30.01        & 1.89         & 21.29           & 3710.5       \\
                        & Validation      & 29.86         & 19.74        & 4.84         & 25.29           & 150.96       \\
                        & Test            & 35.47         & 37.68        & 3.22         & 28.02           & 634.61       \\ \midrule
        EFC             & Overall         & 0.90          & 0.07         & 0.24         & 0.92            & 1.00         \\
                        & Train           & 0.90          & 0.07         & 0.24         & 0.92            & 1.00         \\
                        & Validation      & 0.95          & 0.04         & 0.78         & 0.96            & 1.00         \\
                        & Test            & 0.95          & 0.04         & 0.78         & 0.96            & 1.00         \\ \midrule
        EFD             & Overall         & 4.63          & 4.05         & 0.25         & 3.51            & 77.65        \\
                        & Train           & 4.59          & 3.98         & 0.25         & 3.48            & 77.65        \\
                        & Validation      & 6.90          & 6.31         & 1.76         & 4.80            & 45.56        \\
                        & Test            & 6.02          & 5.49         & 1.58         & 4.54            & 66.40        \\ \midrule
        1GN             & Overall         & 10.15         & 7.85         & 0.00         & 8.55            & 90.38        \\
                        & Train           & 10.26         & 7.89         & 0.00         & 8.70            & 90.38        \\
                        & Validation      & 5.52          & 4.30         & 0.00         & 4.40            & 24.29        \\
                        & Test            & 5.73          & 4.80         & 0.00         & 4.58            & 26.79        \\ \midrule
        2GN             & Overall         & 45.63         & 16.39        & 0.00         & 45.28           & 100.0        \\
                        & Train           & 45.86         & 16.39        & 0.00         & 45.53           & 100.0        \\
                        & Validation      & 36.25         & 13.70        & 7.31         & 37.50           & 76.92        \\
                        & Test            & 37.15         & 13.82        & 9.57         & 36.55           & 76.36        \\ \midrule
        3GN             & Overall         & 64.62         & 17.50        & 0.00         & 66.15           & 100.0        \\
                        & Train           & 64.84         & 17.47        & 0.00         & 66.67           & 100.0        \\
                        & Validation      & 55.38         & 16.87        & 15.06        & 58.49           & 100.0        \\
                        & Test            & 56.95         & 16.25        & 17.65        & 58.14           & 96.30        \\ \midrule
        4GN             & Overall         & 75.46         & 16.86        & 0.00         & 78.43           & 100.0        \\
                        & Train           & 75.65         & 16.82        & 0.00         & 78.65           & 100.0        \\
                        & Validation      & 66.70         & 17.31        & 20.16        & 70.67           & 100.0        \\
                        & Test            & 68.87         & 16.30        & 22.32        & 70.36           & 100.0        \\ \midrule
    \end{tabular}%
    \caption{Summarization-related properties of our dataset for each split. CR = Compression Ratio, EFC/EFD = Extractive Fragment Coverage/Density, 1GN-4GN = n-Gram Novelty percentages. CRs are calculated across all samples, the other metrics only across samples where the decision language matches the headnote language to prevent distorted results due to non-matching n-gram pairs in different languages.}
    \label{tab:summarization_properties}
\end{table*}

\subsection{Connection between language and law area}
\label[appendix]{app:language-and-law-area}

To examine whether language (used here as a proxy for different geographic regions of Switzerland) is associated with the area of law, we conducted a chi-square test of independence. The test indicated a statistically significant association, $\chi^2(14) = 603.67$, $p < .001$, but the effect size was negligible, $V = 0.07$. Thus, for practical purposes, the two variables can be considered independent.

\subsection{Pre-Training Corpora Analysis}
\label[appendix]{app:pretrain-corpora-analysis}

Because the decisions and headnotes in \gls{SLDS} were scraped from the \gls{SFSC} archive, we examined whether they might also appear in widely used pre-training corpora. If so, models could achieve artificially high scores on the \gls{SLDS} test split due to memorization rather than genuine summarization ability. This analysis is challenging, as most \gls{LLM} publishers do not disclose or provide access to their pre-training datasets. We therefore relied on publicly available corpora and performed a membership analysis using the Infini-Gram engine \cite{liu2024infini}, which allows n-gram searches across 11 indices accessible via their API. These indices include well-known corpora such as the Pile \citep{gao2020pile, biderman2022datasheet}, Dolma \citep{dolma}, C4 \citep{raffel2020exploring}, RedPajama \citep{weber2024redpajama}, and DCLM \citep{li2024datacomplm}, as well as corpora used for training OLMo 2 \citep{olmo20242} and OLMoE \citep{muennighoff2024olmoeopenmixtureofexpertslanguage}.

Our procedure was as follows: we randomly sampled 100 decisions from \gls{SLDS} published between 2010 and 2017. From each sample, we extracted nested n-grams of increasing length from identical text positions (e.g., \textit{für die Ausübung des} [n=4], \textit{für die Ausübung des Berufes} [n=5], \textit{für die Ausübung des Berufes erforderlichen} [n=6], etc.).\footnote{We constructed substrings via whitespace tokenization. The Infini-Gram engine applies its own tokenization when processing queries.} We then queried the Infini-Gram API to count occurrences of these n-grams across the selected indices. To reduce unnecessary queries, we applied early stopping: once a shorter n-gram had zero matches, longer n-grams from the same sequence were skipped. For each of the 100 samples, we chose three random starting unigrams, resulting in three sets of nested n-grams per sample. For each set, we analyzed n-grams with $n=4,\dots,10$.

Our results show that 25 out of 300 four-grams were found in at least one pre-training corpus. However, all were either highly generic phrases or contained specific dates or legal references. Among the 300 five-grams, only four occurred in the Infini-Gram indices. For $n \in \{6,7,8\}$, we identified a single overlapping phrase: \textit{"dans la mesure où cela est possible et"} (roughly, \textit{"to the extent possible and"}). Given its generic nature, and the absence of matches among the analyzed nine- and ten-grams, we conclude that it is highly unlikely that \gls{SLDS} was included in any of the analyzed corpora. While we cannot exclude the possibility that other, non-public corpora may contain samples from \gls{SLDS}, their absence from the examined datasets provides reassuring evidence against contamination.

\section{Resources Used}


For fine-tuning and the learning rate sweeps, we mostly used a single NVIDIA H100 GPU with 96 GB of VRAM. Some runs were performed on another node with two NVIDIA A100 GPUs with 80 GB of VRAM each. The total runtime of these experiments was 15.363 days.

\section{Hyperparameters Used in Metrics}
\label[appendix]{app:metric-hyperparams}

For ROUGE, we employed the \texttt{lighteval} wrapper, which internally uses the \texttt{rouge\_score} library with default settings—specifically, whitespace-based tokenization without stemming or additional preprocessing.

For BERTScore, we used \texttt{xlm-roberta-large} through the \texttt{BERTScorer} implementation in \texttt{lighteval}, setting \texttt{rescale\_with\_baseline=True} and \texttt{num\_layers=24}. Language-specific baselines were obtained from the official BERTScore GitHub repository.

\section{Experiment Details}
\label[appendix]{app:experiment-details}

\subsection{Exact Model Versions}

For the proprietary models, we used the following model versions in our experiments: \texttt{gpt-4o-2024-08-06}, \texttt{o3-mini-2025-01-31}, \texttt{claude-3-5-sonnet-20241022}.

\subsection{One-Shot Example Selection Strategy}
To reduce the input sequence length and the associated costs, we selected the sample with the shortest sequence length in the validation split for each decision-headnote language pair and use them as the one-shot examples in our experiment. The one-shot example was provided in terms of a user- and assistant message pair in a multi-turn chat completions format.

\subsection{Decoding Parameters}

We used the default vLLM settings, with some modifications for Llama3.2 3B and the Qwen model family:

\begin{itemize}
    \item \verb|repetition_penalty|: 1.05
    \item \verb|temperature|: 0.7
    \item \verb|top_k|: 20
    \item \verb|top_p|: 0.8
\end{itemize}

We used the seed 2025 for reproducibility. Proprietary model APIs accepted only some of these parameters. More specifically, for OpenAI and DeepSeek models, we had to drop the \verb|repetition_penalty| and the \verb|top_k| parameter. The Anthropic API did not accept a \verb|repetition_penalty| parameter either.

\section{Fine-Tuning Hyperparameters}
\label[appendix]{app:hyperparameters-fine-tuning}

We fine-tuned our models using the Unsloth library \citep{unsloth}. We followed a \gls{PEFT} training scheme by only fine-tuning a small set of additional weights using LoRA \citep{hu2021lora}. We used 16 for both the LoRA rank and the alpha. LoRA was applied to the following target modules: \verb|q_proj|, \verb|k_proj|, \verb|v_proj|, \verb|o_proj|, gat\verb|e_proj|, u\verb|p_proj|, dow\verb|n_proj|. Whenever possible, we used a batch size of 32. Where this was not possible, we used gradient accumulation steps to still train with an effective batch size of 32. For each model, we performed a learning rate sweep across three different learning rates (1e-5, 5e-5, 1e-4) for 500 steps. The 1e-4 learning rate performed best across all models, so we used it for fine-tuning all of our models with 200 warmup steps and a linear learning rate scheduler. We used an 8-bit version of AdamW \citep{loshchilov2017decoupled} as the optimizer and trained the models for 3 epochs. Due to memory limitations, the maximum sequence length of the models was set to 8192, which is long enough to cover roughly 95\% of all decisions in the training set when estimated using the \verb|tiktoken| tokenizer. The rest of the decisions was truncated during training. The exact training configuration along with the training and evaluation scripts can be found on our GitHub repository.

\section{Prompts}
\label{appendix:prompts}

All the models that we used during our experiments use chat templates. Below, we report the different system and user messages that were used in our experiments.

\subsection{Fine-Tuning}

During fine-tuning, we did not specify the system message, which means that the individual default system message for each model was used. The user message that we used to teach the model to map decisions to headnotes was a simple prefix that can be seen below in \Cref{prompt:fine-tuning-user-prompt}.

\begin{figure}[htb]
    \centering
    \begin{framed}
        \begin{minipage}[t]{\textwidth}
            \scriptsize
\begin{Verbatim}
Generate a headnote in \textcolor{blue}{\string{language\string}} for the following leading decision: \textcolor{blue}{\string{decision\string}}
\end{Verbatim}
        \end{minipage}
    \end{framed}
    \captionof{prompt}{The user prompt that was used during fine-tuning. The blue text wrapped with curly brackets represent variables. The decision text was inserted directly from dataset column. For the language, we converted the language ISO code into the corresponding written out language first, i.e. either \textit{German}, \textit{French}, or \textit{Italian.}}
    \label[prompt]{prompt:fine-tuning-user-prompt}
\end{figure}

\subsection{Headnote generation}

\begingroup
\renewcommand*{\figureautorefname}{Prompt}%
During the evaluation, we used \Cref{prompt:headnote-generation-system-prompt} as the system prompt and \Cref{prompt:headnote-generation-user-prompt} as the user message to generate the headnotes. Unlike during fine-tuning, we decided to use a suffix rather than a prefix for the instruction to benefit from prompt caching. In the case of the pre-trained models (OpenAI and Anthropic models as well as DeepSeek R1), we used one-shot prompting as implemented in lighteval: an additional initial turn of conversation is added where the assistant response is already provided with the gold headnote as content.
\endgroup

\begin{figure}[htb]
    \centering
    \begin{framed}
        \begin{minipage}[t]{\textwidth}
            \scriptsize
\begin{Verbatim}
You are a legal expert specializing in Swiss Federal Supreme Court decisions with extensive knowledge of legal terminology and conventions in German, French, and Italian. Your task is to generate a headnote for a provided leading decision. A headnote is a concise summary that captures the key legal points and significance of the decision. It is not merely a summary of the content but highlights the aspects that make the decision "leading" and important for future legislation.

When generating the headnote:

1. Focus on the core legal reasoning and key considerations that establish the decision's significance.
2. Include any relevant references to legal articles (prefixed with "Art.") and considerations (prefixed with "E." in German or "consid." in French/Italian).
3. Use precise legal terminology and adhere to the formal and professional style typical of Swiss Federal Supreme Court headnotes.
4. Ensure clarity and coherence, so the headnote is logically structured and easy to understand in the specified language.

Your response should consist solely of the headnote in the language specified by the user prompt.
\end{Verbatim}
        \end{minipage}
    \end{framed}
    \captionof{prompt}{The system prompt that was used during the generation of the headnotes.}
    \label[prompt]{prompt:headnote-generation-system-prompt}
\end{figure}

\begin{figure}[htb]
    \centering
    \begin{framed}
        \begin{minipage}[t]{\textwidth}
            \scriptsize
\begin{Verbatim}
Leading decision:
```\textcolor{blue}{\string{decision\string}}```

Generate a headnote in \textcolor{blue}{\string{language\string}} for the leading decision above.
\end{Verbatim}
        \end{minipage}
    \end{framed}
    \captionof{prompt}{The user prompt that was used during the generation of the headnotes. The blue text wrapped with curly brackets represent variables. The decision text was inserted directly from dataset column. For the language, we converted the language ISO code into the corresponding written out language first, i.e. either \textit{German}, \textit{French}, or \textit{Italian.}}
    \label[prompt]{prompt:headnote-generation-user-prompt}
\end{figure}

\subsection{Evaluation}
\label[appendix]{app:prompts:judge}

For the LLM-as-a-Judge evaluation, we used \Cref{prompt:judge-system-prompt} as the system message and \Cref{prompt:judge-user-prompt} as the user message. In the user prompt, we provided a one-shot example in German, French or Italian, depending on the language of the generated headnote that was evaluated. For these examples, we use the gold headnotes from the validation set that had the least number of tokens in the respective language. The model generated output in these examples stems from DeepSeek V3 and the scores in these demonstrations were assigned manually. The content of these one-shot examples is presented in  \Cref{example:judge-example-de,example:judge-example-fr,example:judge-example-it}.

\begin{figure}[htb]
    \centering
    \begin{framed}
        \begin{minipage}[t]{\textwidth}
            \scriptsize
\begin{Verbatim}
You are a senior legal expert and quality assurance specialist with over 20 years of experience in Swiss law. You possess native-level proficiency in German, French, and Italian, enabling you to evaluate Swiss Federal Supreme Court headnotes with precision. Your task is to compare the **Official (Gold) Headnote** with a **Model-Generated Headnote** and provide a structured evaluation in five categories. You will carefully analyze each category and provide a short analysis before committing to a score. The categories are:

1. Accuracy & Faithfulness: How well does the Model-Generated Headnote match the essential legal meaning and intent of the Official Headnote?
2. Completeness & Relevance: Does the Model-Generated Headnote include all important points that the Official Headnote emphasizes, without adding irrelevant details?
3. Clarity & Coherence: Is the text well-organized, easy to understand, and coherent in style and structure?
4. Articles: Do the same legal articles (prefixed “Art.”) appear correctly and completely in the Model-Generated Headnote as in the Official Headnote?
5. Considerations: Do the same considerations (prefixed “E.” in German or “consid.” in French/Italian) appear correctly and completely in the Model-Generated Headnote as in the Official Headnote?

For each category, provide a short and concise explanation followed by a score on a scale from 1 to 3:

1: Fails or is substantially flawed.
Major omissions or inaccuracies that fundamentally alter the legal meaning.

2: Largely correct but missing key element(s).
Generally captures the substance, yet lacks one or more important details or references.

3: Closely matches the Official Headnote.
Covers all critical aspects and references with only minor wording variations that do not affect the legal content.

Your output must follow the exact structure provided below to ensure consistency and ease of parsing.
\end{Verbatim}
        \end{minipage}
    \end{framed}
    \captionof{prompt}{The system prompt that was used for the DeepSeek V3 judge in the LLM-as-a-Judge evaluation. It describes the five categories that the judge should use to compare the generated headnotes with the original (gold) headnotes as well as the grading system.}
    \label[prompt]{prompt:judge-system-prompt}
\end{figure}

\begin{figure}[htb]
    \centering
    \begin{framed}
        \begin{minipage}[t]{\textwidth}
            \scriptsize
\begin{Verbatim}
Below are two headnotes for the same leading decision from the Swiss Federal Supreme Court. Please compare the Model-Generated Headnote to the Official (Gold) Headnote according to the following five categories: Accuracy & Faithfulness, Completeness & Relevance, Clarity & Coherence, Articles, and Considerations.

1. Analyze the Model-Generated Headnote in comparison to the Official Headnote for each category.
2. Provide a short explanation for your evaluation in each category.
3. Conclude each category with a score in the exact format: CATEGORYNAME_SCORE: [X], where X is an integer from 1 to 3.

Required Output Format:

ACCURACY_FAITHFULNESS:  
Analysis: [Your concise analysis here]  
ACCURACY_FAITHFULNESS_SCORE: [X]

COMPLETENESS_RELEVANCE:  
Analysis: [Your concise analysis here]  
COMPLETENESS_RELEVANCE_SCORE: [X]

CLARITY_COHERENCE:  
Analysis: [Your concise analysis here]  
CLARITY_COHERENCE_SCORE: [X]

ARTICLES:  
Analysis: [Your concise analysis here]  
ARTICLES_SCORE: [X]

CONSIDERATIONS:  
Analysis: [Your concise analysis here]  
CONSIDERATIONS_SCORE: [X]

Please adhere strictly to this structure.

Here is an example how your response could look like:

<EXAMPLE>
\textcolor{blue}{\string{one_shot_example\string}}
</EXAMPLE>

Now evaluate the Model-Generated Headnote against the Official (Gold) Headnote on the headnotes below.

Official (Gold) Headnote:
\textcolor{blue}{\string{original_headnote\string}}

Model-Generated Headnote:
\textcolor{blue}{\string{generated_headnote\string}}
\end{Verbatim}
        \end{minipage}
    \end{framed}
    \captionof{prompt}{The user prompt that was used for the DeepSeek V3 judge in the LLM-as-a-Judge evaluation. The blue text wrapped with curly brackets represent variables. The \texttt{one\_shot\_example} used either the German, French or Italian demonstration illustrated in \Cref{example:judge-example-de,example:judge-example-fr,example:judge-example-it}}
    \label[prompt]{prompt:judge-user-prompt}
\end{figure}

\begin{figure}[htb]
    \centering
    \begin{framed}
        \begin{minipage}[t]{\textwidth}
            \scriptsize
\begin{Verbatim}
Official (Gold) Headnote:
Art. 391 Abs. 2 StPO; die erstmalige Anordnung einer ambulanten Massnahme durch das Berufungsgericht verstösst gegen das Verschlechterungsverbot (reformatio in peius). Verzichtet das erstinstanzliche Gericht auf die Anordnung einer beantragten ambulanten Massnahme und hat die Staatsanwaltschaft in ihrer Anschlussberufung deren Anordnung nicht erneut beantragt, verletzt das Berufungsgericht das Verschlechterungsverbot, wenn es eine ambulante Massnahme anordnet (E. 4.1-4.4).

Model-Generated Headnote:
Art. 391 Abs. 2 StPO; Verschlechterungsverbot; ambulante Massnahme nach Art. 63 StGB. Die erstmalige Anordnung einer ambulanten Massnahme im Berufungsverfahren verletzt das Verschlechterungsverbot nach Art. 391 Abs. 2 StPO, wenn erstinstanzlich auf eine solche Massnahme verzichtet wurde und die Staatsanwaltschaft in ihrer Anschlussberufung keinen entsprechenden Antrag gestellt hat. Eine solche Anordnung stellt eine unzulässige reformatio in peius dar, da der Beschuldigte das Risiko einer nachträglichen Anpassung oder Umwandlung der Massnahme nur dann trägt, wenn bereits erstinstanzlich eine therapeutische Massnahme angeordnet wurde (E. 4.4). Die freiwillige Teilnahme an einer Therapie durch den Beschuldigten begründet keine ausreichende Grundlage für die erstmalige Anordnung einer ambulanten Massnahme im Berufungsverfahren (E. 4.3).

ACCURACY_FAITHFULNESS:
Analysis: The model-generated headnotes captures the legal essence and intent of the official headnote accurately. It correctly explains the legal principle of reformatio in peius and the violation of the prohibition of worsening in the context of ordering outpatient measures.
ACCURACY_FAITHFULNESS_SCORE: 3

COMPLETENESS_RELEVANCE:
Analysis: The model-generated headnote inclundes all relevant aspects of the official headnote, such as the prohibition of worsening, the legal context of ordering outpatient measures, and the implications of the prosecution's appeal. However, it also adds additional details regarding the voluntary participation in therapy, which are not explicitly mentioned in the official headnote. These are very specific details that may not be necessary for a concise headnote.
COMPLETENESS_RELEVANCE_SCORE: 2

CLARITY_COHERENCE:
Analysis: The model-generated headnote is easy to understand and coherent. The clarity could be improved by not repeating the fact that the outpatient measure violates the prohibition of worsening twice. Other than that, the structure and style are clear and well-organized.
CLARITY_COHERENCE_SCORE: 2

ARTICLES:
Analysis: While both the official and model-generated headnotes correctly reference Art. 391 Abs. 2 StPO, the model-generated headnote also includes Art. 63 StGB, which is not explicitly mentioned in the official headnote. This additional reference is relevant to the context of outpatient measures but may not be necessary for a concise headnote.
ARTICLES_SCORE: 2

CONSIDERATIONS:
Analysis: The model-generated headnote only cites considerations E. 4.3 and E. 4.4, whereas the official headnote includes considerations E. 4.1 to E. 4.4. This omission reduces the completeness of the model-generated headnote.
CONSIDERATIONS_SCORE: 2
\end{Verbatim}
        \end{minipage}
    \end{framed}
    \captionof{example}{The German one-shot example that was used in \Cref{prompt:judge-user-prompt} in the LLM-as-a-Judge evaluation in case the generated and gold headnotes to evaluate were in German. For token efficiency, we used the sample with \texttt{sample\_id} 60312 as it was the shortest possible example from the validation set in German.}
    \label[example]{example:judge-example-de}
\end{figure}

\begin{figure}[htb]
    \centering
    \begin{framed}
        \begin{minipage}[t]{\textwidth}
            \scriptsize
\begin{Verbatim}
Official (Gold) Headnote:
Travail externe (art. 77a CP). Le condamné à une peine privative de liberté qui a passé une longue période en détention provisoire peut exécuter sa peine (restante) directement sous la forme de travail externe s'il réalise les conditions posées à l'art. 77a al. 1 CP. Il n'a pas nécessairement besoin d'être privé de sa liberté au moment où est prise la décision ordonnant le travail externe (consid. 2.5.2).

Model-Generated Headnote:
Art. 77a CP; art. 164 ss RSPC; conditions pour l'octroi du travail externe. Le travail externe, en tant que phase de l'élargissement progressif de l'exécution de la peine, ne nécessite pas impérativement un séjour préalable en milieu ouvert ou dans une section ouverte d'un établissement fermé. L'utilisation du terme "en principe" à l'art. 77a al. 2 CP indique que cette condition n'est pas absolue et peut être adaptée aux circonstances individuelles, notamment lorsque le condamné a déjà purgé une partie significative de sa peine en détention provisoire ou pour des motifs de sûreté. La cour cantonale a violé le droit fédéral en exigeant de manière rigide un séjour en milieu ouvert comme condition préalable à l'octroi du travail externe. Le recours est admis, l'arrêt cantonal est annulé, et la cause est renvoyée pour une nouvelle décision sur la base des conditions prévues à l'art. 77a CP (consid. 2.5.1 à 2.5.3).

ACCURACY_FAITHFULNESS:  
Analysis: The model-generated headnote accurately reflects the legal principle and conditions for granting external work under Art. 77a CP.
ACCURACY_FAITHFULNESS_SCORE: 3

COMPLETENESS_RELEVANCE:  
Analysis: The model-generated headnote includes all relevant aspects of the official headnote. However, it adds additional details regarding the use of the term "en principe" and the violation of federal law by the cantonal court. While these details provide context, they are not essential for a concise headnote that shapes future legislation.
COMPLETENESS_RELEVANCE_SCORE: 2

CLARITY_COHERENCE:  
Analysis: The model-generated headnote is clear and well-organized, but the inclusion of specific details may obscure the broader legal principle.
CLARITY_COHERENCE_SCORE: 2

ARTICLES:  
Analysis: The model-generated headnote includes extra legal articles (Art. 164 ff. RSPC) not cited in the official headnote, deviating from its intended focus. Besides this, the reference to Art. 77a CP aligns with the official headnote.
ARTICLES_SCORE: 2

CONSIDERATIONS:  
Analysis: The official headnote only cites consideration E 2.5.2, while the model-generated headnote references considerations E 2.5.1 to E 2.5.3. This makes the model-generated headnote less precise.
CONSIDERATIONS_SCORE: 2
\end{Verbatim}
        \end{minipage}
    \end{framed}
    \captionof{example}{The French one-shot example that was used in \Cref{prompt:judge-user-prompt} in the LLM-as-a-Judge evaluation in case the generated and gold headnotes to evaluate were in French. For token efficiency, we used the sample with \texttt{sample\_id} 60247 as it was the shortest possible example from the validation set in German.}
    \label[example]{example:judge-example-fr}
\end{figure}

\begin{figure}[htb]
    \centering
    \begin{framed}
        \begin{minipage}[t]{\textwidth}
            \scriptsize
\begin{Verbatim}
Official (Gold) Headnote:
Art. 24 Cost.; libertà di domicilio; residenza secondaria per motivi professionali; autorizzazione di soggiorno.  Contenuto e portata della libertà di domicilio garantita dall'art. 24 Cost. (consid. 3.2.1). In tale ambito le competenze dei Cantoni e dei Comuni sono circoscritte al controllo e alla registrazione degli abitanti (consid. 3.3.1). L'obbligo di annunciarsi, in caso di soggiorno in un Comune differente da quello del domicilio per motivi professionali, non limita l'esercizio della libertà di domicilio. Detto annuncio non può però essere sottoposto ad un regime autorizzativo (consid. 3.3.3).

Model-Generated Headnote:
Art. 24 Cost.; libertà di domicilio; regime autorizzativo per il soggiorno; competenze comunali. La libertà di domicilio garantita dall'art. 24 Cost. consente a ogni cittadino svizzero di eleggere liberamente il proprio luogo di residenza o soggiorno sul territorio nazionale, senza che ciò possa essere sottoposto a un regime autorizzativo di polizia. Le autorità comunali possono richiedere unicamente una notifica al momento dell'arrivo nel Comune, finalizzata alla registrazione e al controllo degli abitanti, ma non possono imporre un'autorizzazione di soggiorno con scadenza annuale, poiché ciò costituirebbe una violazione della libertà di domicilio (consid. 3.3.3). Il sistema instaurato dalle Direttive MOVPOP, che prevede il rilascio di un'"autorizzazione di soggiorno" con validità limitata, deve essere interpretato nel senso che l'autorità comunale può solo certificare formalmente la notifica del soggiorno, senza sottoporre quest'ultimo a un regime autorizzativo (consid. 3.3.2 e 3.3.3). La conferma di un tale regime da parte del Tribunale cantonale amministrativo viola pertanto l'art. 24 Cost. e deve essere annullata (consid. 3.4).

ACCURACY_FAITHFULNESS:  
Analysis: The model-generated headnote aligns with the core legal meaning but includes additional details (e.g., MOVPOP directives) not in the official headnote. These do not conflict but shift the focus slightly.
ACCURACY_FAITHFULNESS_SCORE: 2

COMPLETENESS_RELEVANCE:  
Analysis: The model-generated headnote captures key points but omits emphasis on secondary residence for professional reasons and cantonal/communal roles. Irrelevant details (e.g., MOVPOP) add complexity. 
COMPLETENESS_RELEVANCE_SCORE: 2

CLARITY_COHERENCE:  
Analysis: The model-generated headnote is clear and organized, but additional elements like MOVPOP reduce coherence by shifting focus away from the main points and making the text longer and more complex.
CLARITY_COHERENCE_SCORE: 2

ARTICLES:  
Analysis: References to Art. 24 Cost. are correct and complete.
ARTICLES_SCORE: 3

CONSIDERATIONS:  
Analysis: The model-generated headnote correctly references consid. 3.3.3 but adds consid. 3.3.2 and 3.4, which are beyond the official headnote's scope. Moreover, it leaves out consid 3.2.1 and 3.3.1, reducing precision. Instead, it mentiones consid. 3.3.3 twice, which is redundant.
CONSIDERATIONS_SCORE: 1
\end{Verbatim}
        \end{minipage}
    \end{framed}
    \captionof{example}{The Italian one-shot example that was used in \Cref{prompt:judge-user-prompt} in the LLM-as-a-Judge evaluation in case the generated and gold headnotes to evaluate were in Italian. For token efficiency, we used the sample with \texttt{sample\_id} 59894 as it was the shortest possible example from the validation set in German.}
    \label[example]{example:judge-example-it}
\end{figure}

\section{Judge Example Output}
\label[appendix]{app:examples:judge}

An example output of the DeepSeek V3 judge below can be seen in \Cref{fig:judge-example-output}.

\begin{figure*}[tb]
    \vspace{-2ex}
    \begin{judgeExample}
        {61194}
        {Claude 3.5 Sonnet}
        {40}
        {\input{examples/judge/61194/generated_headnote}}
        {\input{examples/judge/61194/generated_headnote_translated}}
        {\input{examples/judge/61194/original_headnote}}
        {\input{examples/judge/61194/original_headnote_translated}}
        {\input{examples/judge/61194/judge_output}}
    \end{judgeExample}
    \vspace{-2ex}
    \begin{adjustwidth}{-1cm}{-1cm}
        \caption{Example of a judgment made by the DeepSeek V3 judge for a headnote generated for sample with ID 61194 using Claude 3.5 Sonnet. Given the five scores that the judge produces across the five evaluation category, an aggregated score of 40 is computed.}
        \label{fig:judge-example-output}
    \end{adjustwidth}
\end{figure*}

\section{Output Artefacts and Score Correction for Claude 3.5 Sonnet}
\label[appendix]{app:claude-negative-bertscores}

During our evaluation of Claude 3.5 Sonnet, we observed two distinct output behaviors that required special handling to ensure a fair and accurate assessment. These artefacts, empty outputs and unconventional formatting, and the corresponding corrections are detailed below.

\subsection{Correction of Negative BERTScores from Empty Outputs}

The initial aggregated BERTScore results for Claude 3.5 Sonnet, as calculated by \texttt{lighteval}, were unreasonably negative. Our investigation revealed that this was caused by four samples where the model produced an empty string as output.\footnote{It was actually an empty JSON, i.e. \texttt{"\{\}"}, but the opening and closing bracket were removed in a pre-processing step in \texttt{lighteval} before computing the BERTScore.} The BERTScore implementation assigned these empty outputs extremely low negative scores (in the range of -5000), which disproportionately skewed the overall average.

To rectify this, we replaced these four outlier scores with a score of zero. This correction is logically sound, as a zero score accurately represents a complete lack of overlapping content between a null output and the reference text. All final BERTScore results for Claude 3.5 Sonnet reported in this work have been calculated with this correction applied.

\subsection{Output Format}

A second notable behaviour was the model's tendency to wrap its generated headnotes in a JSON object, even though this format was not requested in the prompt. The output typically contained the headnote as a value associated with a key such as \textit{headnote}, \textit{text}, or \textit{input}. Additionally, the text within these JSON outputs often contained Unicode-escaped characters for German umlauts (e.g., \verb|\u00e4| instead of \verb|ä|).

We hypothesize that this combination of JSON syntax and character escaping explains the observed discrepancy between the model's relatively low BERTScore and its high JUDGE score:

\begin{itemize}
    \item \textbf{BERTScore:} This metric measures token-level similarity. The presence of extraneous JSON characters and escaped character sequences introduced tokens that did not match the plain-text reference, thereby penalizing the score.
    \item \textbf{JUDGE (LLM-based):} In contrast, the LLM judge was capable of parsing the JSON structure and correctly interpreting the Unicode-escaped characters. It could therefore "look past" the formatting and evaluate the semantic quality of the underlying headnote accurately, resulting in a higher score.
\end{itemize}

\section{Results on Language Subsets}
\label[appendix]{app:results-cross-lingual}

We provide the detailed results for the cross-lingual evaluations in our experiment in \Cref{tab:results-cross-lingual}.

\clearpage
\onecolumn

{\scriptsize
    \begin{xltabular}{\textwidth}{|l|l|l|X|X|X|X|X|X|}
        \toprule
        \textbf{Model} & \textbf{Subset} & \textbf{BERTScore} $\uparrow$ & \textbf{BLEU} $\uparrow$ & \textbf{ROUGE-1} $\uparrow$ & \textbf{ROUGE-2} $\uparrow$ & \textbf{ROUGE-L} $\uparrow$ & \textbf{JUDGE} $\uparrow$ \\
        \midrule
        \endfirsthead
        \toprule
        \textbf{Model} & \textbf{Subset} & \textbf{BERTScore} $\uparrow$ & \textbf{BLEU} $\uparrow$ & \textbf{ROUGE-1} $\uparrow$ & \textbf{ROUGE-2} $\uparrow$ & \textbf{ROUGE-L} $\uparrow$ & \textbf{JUDGE} $\uparrow$ \\
        \midrule
        \endhead
        Phi-3.5-mini & de\_de & 6.74 ± 2.63 & 40.34 ± 0.54 & 31.30 ± 1.19 & 15.65 ± 0.91 & 22.76 ± 0.94 & 20.97 ± 1.55 \\
        Llama 3.2 3B & de\_de & 27.27 ± 1.43 & 47.59 ± 0.45 & 39.58 ± 1.02 & 20.72 ± 0.99 & 28.84 ± 0.94 & 28.50 ± 1.56 \\
        Qwen2.5 0.5B & de\_de & 16.37 ± 1.65 & 72.53 ± 0.41 & 31.51 ± 0.87 & 14.45 ± 0.64 & 23.08 ± 0.67 & 13.00 ± 1.15 \\
        Qwen2.5 1.5B & de\_de & 23.19 ± 1.49 & \textbf{74.22 ± 0.44} & 36.05 ± 0.92 & 17.72 ± 0.84 & 26.37 ± 0.86 & 21.88 ± 1.38 \\
        Qwen2.5 3B & de\_de & 28.22 ± 1.40 & 67.40 ± 0.41 & 39.31 ± 0.93 & 20.20 ± 0.88 & 29.10 ± 0.87 & 29.42 ± 1.62 \\
        Qwen2.5 7B & de\_de & 32.21 ± 1.24 & 72.18 ± 0.43 & 42.26 ± 0.98 & 22.78 ± 1.06 & 31.64 ± 1.01 & 33.09 ± 1.50 \\
        Qwen2.5 14B & de\_de & \textbf{35.22 ± 1.22} & 66.74 ± 0.43 & \textbf{43.82 ± 0.94} & \textbf{24.54 ± 1.08} & \textbf{33.48 ± 1.02} & 36.47 ± 1.60 \\
        GPT-4o & de\_de & 27.96 ± 0.90 & 39.94 ± 0.26 & 40.78 ± 0.69 & 18.72 ± 0.62 & 26.97 ± 0.56 & 40.58 ± 1.33 \\
        DeepSeek-R1 & de\_de & 17.29 ± 0.74 & 29.69 ± 0.19 & 36.04 ± 0.64 & 15.01 ± 0.44 & 21.02 ± 0.38 & \textbf{43.91 ± 1.16} \\
        o3-mini & de\_de & 13.78 ± 0.73 & 31.34 ± 0.20 & 33.04 ± 0.54 & 11.85 ± 0.40 & 18.18 ± 0.34 & 36.52 ± 1.09 \\
        Claude 3.5 Sonnet & de\_de & 0.53 ± 1.28 & 27.00 ± 0.28 & 40.50 ± 0.77 & 18.66 ± 0.72 & 29.24 ± 0.76 & 42.27 ± 1.41 \\
        \midrule\midrule
        Phi-3.5-mini & de\_fr & 4.71 ± 2.47 & 50.73 ± 0.52 & 27.36 ± 1.18 & 11.48 ± 0.63 & 18.50 ± 0.76 & 13.57 ± 1.33 \\
        Llama 3.2 3B & de\_fr & 24.84 ± 1.62 & 18.07 ± 0.41 & 35.29 ± 0.92 & 15.16 ± 0.53 & 24.03 ± 0.62 & 19.08 ± 1.40 \\
        Qwen2.5 0.5B & de\_fr & -3.81 ± 2.18 & 20.30 ± 0.50 & 22.33 ± 0.92 & 7.23 ± 0.42 & 15.77 ± 0.55 & 3.29 ± 0.48 \\
        Qwen2.5 1.5B & de\_fr & 21.71 ± 1.61 & 25.19 ± 0.38 & 33.69 ± 0.87 & 13.08 ± 0.53 & 22.28 ± 0.55 & 11.79 ± 1.09 \\
        Qwen2.5 3B & de\_fr & 26.37 ± 1.32 & 40.22 ± 0.32 & 35.87 ± 0.76 & 14.39 ± 0.48 & 24.06 ± 0.50 & 18.55 ± 1.29 \\
        Qwen2.5 7B & de\_fr & 32.61 ± 1.06 & \textbf{52.55 ± 0.32} & 40.56 ± 0.74 & 17.94 ± 0.59 & 26.69 ± 0.56 & 26.47 ± 1.52 \\
        Qwen2.5 14B & de\_fr & 33.78 ± 1.15 & 40.47 ± 0.41 & 40.67 ± 0.80 & 19.44 ± 0.63 & 28.30 ± 0.63 & 30.92 ± 1.55 \\
        GPT-4o & de\_fr & \textbf{33.97 ± 0.76} & 30.45 ± 0.21 & \textbf{45.47 ± 0.61} & \textbf{20.65 ± 0.50} & 27.59 ± 0.42 & 40.14 ± 1.42 \\
        DeepSeek-R1 & de\_fr & 20.84 ± 0.61 & 24.25 ± 0.15 & 39.69 ± 0.66 & 16.36 ± 0.37 & 21.49 ± 0.29 & 42.80 ± 1.24 \\
        o3-mini & de\_fr & 15.68 ± 0.62 & 20.86 ± 0.15 & 36.99 ± 0.56 & 13.11 ± 0.31 & 18.78 ± 0.25 & 35.70 ± 1.33 \\
        Claude 3.5 Sonnet & de\_fr & -5.74 ± 0.94 & 27.23 ± 0.21 & 43.15 ± 0.64 & 19.34 ± 0.55 & \textbf{28.88 ± 0.56} & \textbf{44.88 ± 1.48} \\
        \midrule\midrule
        Phi-3.5-mini & de\_it & 8.06 ± 2.28 & 30.39 ± 0.47 & 25.85 ± 1.03 & 9.89 ± 0.52 & 18.08 ± 0.72 & 9.61 ± 1.09 \\
        Llama 3.2 3B & de\_it & 22.81 ± 1.60 & 14.32 ± 0.41 & 31.47 ± 0.78 & 12.65 ± 0.50 & 22.46 ± 0.56 & 13.72 ± 1.28 \\
        Qwen2.5 0.5B & de\_it & 4.48 ± 1.89 & \textbf{48.16 ± 0.38} & 22.19 ± 0.76 & 7.62 ± 0.34 & 16.35 ± 0.50 & 2.17 ± 0.40 \\
        Qwen2.5 1.5B & de\_it & 22.99 ± 1.30 & 41.46 ± 0.33 & 30.71 ± 0.69 & 10.86 ± 0.42 & 21.31 ± 0.50 & 8.16 ± 0.88 \\
        Qwen2.5 3B & de\_it & 23.86 ± 1.50 & 31.39 ± 0.33 & 32.30 ± 0.77 & 12.41 ± 0.47 & 22.53 ± 0.56 & 12.46 ± 1.24 \\
        Qwen2.5 7B & de\_it & 30.75 ± 1.00 & 31.86 ± 0.34 & 35.74 ± 0.71 & 14.77 ± 0.53 & 24.99 ± 0.56 & 20.39 ± 1.44 \\
        Qwen2.5 14B & de\_it & \textbf{34.46 ± 0.95} & 45.34 ± 0.35 & 37.86 ± 0.68 & 16.38 ± 0.54 & 26.72 ± 0.54 & 25.12 ± 1.44 \\
        GPT-4o & de\_it & 32.12 ± 0.69 & 30.40 ± 0.25 & 39.05 ± 0.58 & 15.51 ± 0.49 & 24.78 ± 0.44 & 29.66 ± 1.29 \\
        DeepSeek-R1 & de\_it & 23.98 ± 0.55 & 12.77 ± 0.17 & 36.07 ± 0.53 & 12.72 ± 0.34 & 20.15 ± 0.29 & 30.14 ± 1.26 \\
        o3-mini & de\_it & 15.90 ± 0.52 & 15.63 ± 0.14 & 30.70 ± 0.45 & 7.79 ± 0.24 & 15.58 ± 0.23 & 27.83 ± 1.23 \\
        Claude 3.5 Sonnet & de\_it & 29.59 ± 0.88 & 29.52 ± 0.26 & \textbf{43.26 ± 0.67} & \textbf{20.46 ± 0.63} & \textbf{29.56 ± 0.62} & \textbf{36.52 ± 1.46} \\
        \midrule\midrule
        Phi-3.5-mini & fr\_de & -6.11 ± 3.27 & 38.47 ± 0.41 & 24.14 ± 1.27 & 8.92 ± 0.61 & 16.55 ± 0.82 & 8.69 ± 1.56 \\
        Llama 3.2 3B & fr\_de & 1.58 ± 2.44 & 49.67 ± 0.37 & 25.75 ± 1.11 & 10.72 ± 0.67 & 19.26 ± 0.83 & 10.65 ± 1.56 \\
        Qwen2.5 0.5B & fr\_de & -10.66 ± 2.47 & 33.38 ± 0.39 & 21.14 ± 0.95 & 6.93 ± 0.51 & 15.61 ± 0.64 & 2.71 ± 0.60 \\
        Qwen2.5 1.5B & fr\_de & 0.62 ± 2.21 & 27.16 ± 0.35 & 26.46 ± 0.97 & 9.37 ± 0.57 & 18.88 ± 0.65 & 7.10 ± 1.18 \\
        Qwen2.5 3B & fr\_de & 7.68 ± 2.03 & 28.04 ± 0.32 & 28.78 ± 0.96 & 10.87 ± 0.58 & 20.36 ± 0.65 & 13.36 ± 1.48 \\
        Qwen2.5 7B & fr\_de & 15.63 ± 1.80 & \textbf{50.67 ± 0.31} & 33.45 ± 0.91 & 12.38 ± 0.63 & 22.58 ± 0.62 & 22.90 ± 2.01 \\
        Qwen2.5 14B & fr\_de & \textbf{21.84 ± 1.51} & 41.26 ± 0.34 & 35.59 ± 0.90 & 14.74 ± 0.71 & 24.70 ± 0.66 & 30.65 ± 1.97 \\
        GPT-4o & fr\_de & 21.02 ± 1.03 & 31.29 ± 0.21 & \textbf{38.97 ± 0.72} & \textbf{15.74 ± 0.54} & 24.48 ± 0.50 & 41.12 ± 1.64 \\
        DeepSeek-R1 & fr\_de & 8.17 ± 1.01 & 20.77 ± 0.17 & 33.01 ± 0.70 & 12.30 ± 0.42 & 19.03 ± 0.41 & \textbf{43.64 ± 1.40} \\
        o3-mini & fr\_de & 0.81 ± 0.88 & 19.15 ± 0.18 & 28.94 ± 0.56 & 7.89 ± 0.34 & 15.78 ± 0.33 & 28.69 ± 1.72 \\
        Claude 3.5 Sonnet & fr\_de & -6.30 ± 1.50 & 0.00 ± 0.26 & 36.83 ± 0.74 & 14.80 ± 0.60 & \textbf{24.78 ± 0.60} & 42.90 ± 1.93 \\
        \midrule\midrule
        Phi-3.5-mini & fr\_fr & 18.62 ± 3.27 & 49.91 ± 0.54 & 36.72 ± 1.64 & 18.45 ± 1.22 & 24.61 ± 1.15 & 24.58 ± 2.09 \\
        Llama 3.2 3B & fr\_fr & 24.86 ± 3.03 & 4.32 ± 0.61 & 39.08 ± 1.83 & 21.49 ± 1.42 & 26.75 ± 1.30 & 33.36 ± 2.22 \\
        Qwen2.5 0.5B & fr\_fr & 14.65 ± 3.22 & \textbf{51.91 ± 0.50} & 32.02 ± 1.59 & 15.80 ± 1.08 & 22.12 ± 1.03 & 14.30 ± 1.81 \\
        Qwen2.5 1.5B & fr\_fr & 33.37 ± 2.17 & 41.51 ± 0.47 & 42.66 ± 1.35 & 23.66 ± 1.09 & 29.17 ± 1.04 & 31.50 ± 1.92 \\
        Qwen2.5 3B & fr\_fr & 34.57 ± 2.18 & 47.78 ± 0.41 & 44.14 ± 1.37 & 24.20 ± 1.18 & 30.24 ± 1.13 & 35.42 ± 1.93 \\
        Qwen2.5 7B & fr\_fr & 39.91 ± 1.48 & 51.20 ± 0.42 & 47.91 ± 1.08 & 26.80 ± 1.04 & 32.55 ± 0.93 & 38.97 ± 1.90 \\
        Qwen2.5 14B & fr\_fr & \textbf{43.31 ± 1.26} & 42.67 ± 0.44 & 50.06 ± 1.10 & \textbf{29.13 ± 1.17} & \textbf{34.69 ± 1.03} & 41.96 ± 1.99 \\
        GPT-4o & fr\_fr & 40.20 ± 0.96 & 44.32 ± 0.28 & \textbf{50.66 ± 0.81} & 26.53 ± 0.83 & 31.05 ± 0.69 & 48.04 ± 1.48 \\
        DeepSeek-R1 & fr\_fr & 28.07 ± 0.85 & 31.18 ± 0.20 & 43.28 ± 0.93 & 21.53 ± 0.61 & 23.95 ± 0.50 & 49.25 ± 1.38 \\
        o3-mini & fr\_fr & 25.92 ± 0.86 & 34.85 ± 0.21 & 44.01 ± 0.82 & 20.09 ± 0.60 & 23.58 ± 0.45 & 43.93 ± 1.47 \\
        Claude 3.5 Sonnet & fr\_fr & -3.96 ± 1.24 & 17.32 ± 0.24 & 46.57 ± 0.85 & 22.12 ± 0.75 & 30.57 ± 0.76 & \textbf{50.00 ± 1.99} \\
        \midrule\midrule
        Phi-3.5-mini & fr\_it & 17.03 ± 2.96 & 25.76 ± 0.47 & 31.07 ± 1.43 & 12.63 ± 0.77 & 20.79 ± 0.94 & 13.18 ± 1.62 \\
        Llama 3.2 3B & fr\_it & 22.19 ± 2.42 & 4.98 ± 0.47 & 32.31 ± 1.32 & 14.29 ± 0.87 & 22.77 ± 0.95 & 17.57 ± 1.82 \\
        Qwen2.5 0.5B & fr\_it & 5.93 ± 2.73 & 21.94 ± 0.37 & 24.88 ± 1.15 & 9.53 ± 0.64 & 17.93 ± 0.76 & 3.36 ± 0.70 \\
        Qwen2.5 1.5B & fr\_it & 26.50 ± 1.77 & 38.52 ± 0.34 & 34.46 ± 0.92 & 13.10 ± 0.66 & 22.93 ± 0.68 & 12.80 ± 1.34 \\
        Qwen2.5 3B & fr\_it & 28.52 ± 1.93 & 39.51 ± 0.34 & 35.37 ± 1.08 & 15.02 ± 0.76 & 24.62 ± 0.84 & 17.76 ± 1.82 \\
        Qwen2.5 7B & fr\_it & 31.50 ± 1.79 & \textbf{45.05 ± 0.31} & 37.51 ± 1.17 & 16.43 ± 0.79 & 25.69 ± 0.80 & 24.30 ± 2.04 \\
        Qwen2.5 14B & fr\_it & 35.45 ± 1.53 & 44.31 ± 0.33 & 40.03 ± 1.17 & 19.37 ± 0.92 & 28.54 ± 0.95 & 30.65 ± 1.98 \\
        GPT-4o & fr\_it & \textbf{36.37 ± 1.01} & 31.56 ± 0.25 & 42.97 ± 0.79 & 18.84 ± 0.66 & 26.81 ± 0.65 & 32.71 ± 1.66 \\
        DeepSeek-R1 & fr\_it & 26.76 ± 0.91 & 21.21 ± 0.17 & 38.08 ± 0.86 & 15.46 ± 0.54 & 21.31 ± 0.48 & \textbf{38.22 ± 1.66} \\
        o3-mini & fr\_it & 22.98 ± 0.88 & 15.31 ± 0.19 & 36.12 ± 0.65 & 11.22 ± 0.41 & 19.34 ± 0.43 & 29.91 ± 1.60 \\
        Claude 3.5 Sonnet & fr\_it & 29.76 ± 1.25 & 24.62 ± 0.29 & \textbf{45.12 ± 0.96} & \textbf{22.30 ± 0.84} & \textbf{30.11 ± 0.79} & 37.20 ± 1.80 \\
        \midrule\midrule
        Phi-3.5-mini & it\_de & 0.53 ± 6.69 & 20.35 ± 0.23 & 27.05 ± 3.61 & 10.75 ± 1.89 & 17.19 ± 1.69 & 5.83 ± 2.60 \\
        Llama 3.2 3B & it\_de & -3.89 ± 5.97 & 15.89 ± 0.21 & 24.22 ± 3.08 & 10.13 ± 1.91 & 17.67 ± 2.38 & 7.50 ± 3.92 \\
        Qwen2.5 0.5B & it\_de & -23.28 ± 5.94 & 9.64 ± 0.18 & 16.15 ± 2.65 & 5.97 ± 1.11 & 12.09 ± 1.66 & 0.00 ± 0.00 \\
        Qwen2.5 1.5B & it\_de & 4.91 ± 2.90 & 15.66 ± 0.23 & 27.51 ± 2.49 & 9.62 ± 1.36 & 18.77 ± 1.52 & 4.17 ± 2.29 \\
        Qwen2.5 3B & it\_de & 4.32 ± 5.98 & 10.03 ± 0.26 & 28.31 ± 3.07 & 9.06 ± 1.41 & 18.70 ± 1.71 & 10.83 ± 3.36 \\
        Qwen2.5 7B & it\_de & 14.69 ± 3.46 & 21.69 ± 0.27 & 33.39 ± 2.81 & 12.95 ± 2.20 & 21.07 ± 1.92 & 23.33 ± 6.20 \\
        Qwen2.5 14B & it\_de & \textbf{17.83 ± 3.40} & \textbf{28.24 ± 0.36} & 31.46 ± 2.54 & 14.68 ± 2.07 & 22.35 ± 2.15 & 27.50 ± 6.17 \\
        GPT-4o & it\_de & 14.71 ± 2.94 & 21.30 ± 0.20 & 34.98 ± 3.34 & 14.19 ± 1.76 & 21.21 ± 1.82 & 41.67 ± 5.34 \\
        DeepSeek-R1 & it\_de & 5.76 ± 2.42 & 22.03 ± 0.18 & 35.15 ± 3.76 & 13.41 ± 1.41 & 17.94 ± 1.55 & \textbf{45.00 ± 3.99} \\
        o3-mini & it\_de & -6.59 ± 1.74 & 5.54 ± 0.13 & 25.97 ± 2.53 & 6.71 ± 0.68 & 13.16 ± 0.93 & 34.17 ± 3.79 \\
        Claude 3.5 Sonnet & it\_de & -10.25 ± 3.24 & 22.41 ± 0.20 & \textbf{37.18 ± 2.77} & \textbf{14.86 ± 1.53} & \textbf{23.24 ± 2.04} & 40.83 ± 5.29 \\
        \midrule\midrule
        Phi-3.5-mini & it\_fr & 15.30 ± 8.17 & 30.01 ± 0.32 & 33.66 ± 4.87 & 15.59 ± 2.84 & 21.46 ± 3.02 & 13.33 ± 3.76 \\
        Llama 3.2 3B & it\_fr & 11.77 ± 9.72 & 9.48 ± 0.36 & 31.36 ± 5.09 & 14.07 ± 3.08 & 20.35 ± 3.01 & 17.50 ± 6.64 \\
        Qwen2.5 0.5B & it\_fr & -23.29 ± 6.14 & 8.88 ± 0.18 & 17.07 ± 3.10 & 5.40 ± 1.45 & 12.95 ± 1.66 & 9.17 ± 3.36 \\
        Qwen2.5 1.5B & it\_fr & 20.02 ± 5.31 & 24.91 ± 0.22 & 32.04 ± 3.87 & 13.53 ± 2.01 & 20.87 ± 1.87 & 17.50 ± 4.63 \\
        Qwen2.5 3B & it\_fr & 27.60 ± 3.78 & \textbf{39.09 ± 0.32} & 36.43 ± 3.68 & 15.66 ± 2.70 & 22.57 ± 1.85 & 25.00 ± 5.71 \\
        Qwen2.5 7B & it\_fr & 31.67 ± 2.34 & 23.05 ± 0.24 & 39.93 ± 2.92 & 19.09 ± 1.84 & 25.36 ± 1.55 & 34.17 ± 4.99 \\
        Qwen2.5 14B & it\_fr & 31.69 ± 3.27 & 35.41 ± 0.28 & 37.40 ± 3.25 & 16.76 ± 2.31 & 22.95 ± 1.45 & 30.83 ± 7.12 \\
        GPT-4o & it\_fr & \textbf{33.10 ± 3.64} & 31.58 ± 0.23 & \textbf{45.76 ± 4.22} & \textbf{20.92 ± 2.48} & \textbf{26.60 ± 1.98} & 43.33 ± 4.66 \\
        DeepSeek-R1 & it\_fr & 23.65 ± 3.24 & 19.29 ± 0.19 & 43.50 ± 4.22 & 19.51 ± 1.87 & 22.92 ± 1.75 & \textbf{48.33 ± 4.41} \\
        o3-mini & it\_fr & 17.25 ± 3.07 & 16.06 ± 0.14 & 39.77 ± 3.90 & 13.73 ± 1.77 & 20.26 ± 1.57 & 38.33 ± 4.41 \\
        Claude 3.5 Sonnet & it\_fr & -8.72 ± 3.58 & 19.08 ± 0.23 & 42.18 ± 3.96 & 18.76 ± 2.71 & 25.64 ± 2.67 & 40.00 ± 5.50 \\
        \midrule\midrule
        Phi-3.5-mini & it\_it & 36.33 ± 2.62 & 27.64 ± 0.21 & \textbf{43.65 ± 2.48} & 23.63 ± 2.08 & 28.72 ± 2.12 & 27.50 ± 5.24 \\
        Llama 3.2 3B & it\_it & 5.40 ± 11.34 & 32.69 ± 0.52 & 27.97 ± 5.91 & 14.61 ± 4.54 & 20.23 ± 3.83 & 18.33 ± 6.49 \\
        Qwen2.5 0.5B & it\_it & 7.31 ± 8.42 & 23.08 ± 0.28 & 27.58 ± 3.09 & 12.25 ± 2.32 & 20.48 ± 2.37 & 4.17 ± 2.88 \\
        Qwen2.5 1.5B & it\_it & 24.95 ± 5.68 & \textbf{42.49 ± 0.35} & 33.68 ± 3.47 & 16.30 ± 2.21 & 23.44 ± 2.50 & 28.33 ± 5.75 \\
        Qwen2.5 3B & it\_it & 27.92 ± 5.05 & 42.30 ± 0.34 & 36.14 ± 3.46 & 19.11 ± 2.62 & 24.70 ± 2.43 & 20.00 ± 5.50 \\
        Qwen2.5 7B & it\_it & 37.34 ± 3.52 & 24.37 ± 0.41 & 42.38 ± 2.96 & 21.22 ± 2.61 & 27.41 ± 2.40 & 31.67 ± 6.01 \\
        Qwen2.5 14B & it\_it & \textbf{38.77 ± 3.58} & 31.79 ± 0.36 & 43.45 ± 4.30 & \textbf{24.88 ± 3.30} & \textbf{30.33 ± 3.13} & 28.33 ± 4.90 \\
        GPT-4o & it\_it & 34.48 ± 3.73 & 26.14 ± 0.34 & 40.44 ± 4.33 & 19.15 ± 3.11 & 23.81 ± 2.41 & \textbf{40.00 ± 5.08} \\
        DeepSeek-R1 & it\_it & 27.97 ± 2.70 & 20.12 ± 0.19 & 39.91 ± 4.11 & 17.47 ± 1.71 & 21.47 ± 1.91 & 39.17 ± 3.36 \\
        o3-mini & it\_it & 21.87 ± 2.50 & 26.18 ± 0.17 & 37.37 ± 2.83 & 14.92 ± 1.48 & 19.27 ± 1.49 & 38.33 ± 5.05 \\
        Claude 3.5 Sonnet & it\_it & 24.84 ± 4.07 & 29.71 ± 0.29 & 41.98 ± 3.36 & 21.75 ± 2.39 & 27.05 ± 2.05 & 36.67 ± 5.27 \\
        \midrule\midrule
        \bottomrule
        \caption{Results of the baseline experiments on different subsets of the test set of \gls{SLDS}. Each subset is a combination of the decision language and the headnote language. Standard errors are estimated using the bootstrapping mechanism implemented in lighteval \citep{lighteval}. The Phi-3.5-mini, Llama 3.2 and Qwen 2.5 models were fine-tuned and evaluated in a zero-shot manner, the other models were not fine-tuned and evaluated in a one-shot setting. ROUGE scores are multiplied by 100 for readability. JUDGE = LLM as Judge. \textbf{Bold}: best within subset.} \label{tab:results-cross-lingual}
    \end{xltabular}
}

\clearpage
\twocolumn

\section{Off-the-Shelf Performance}
\label[appendix]{app:off-the-shelf-vs-finetuned}

\begin{figure}[htb]
    \centering
    \includegraphics[width=1\linewidth]{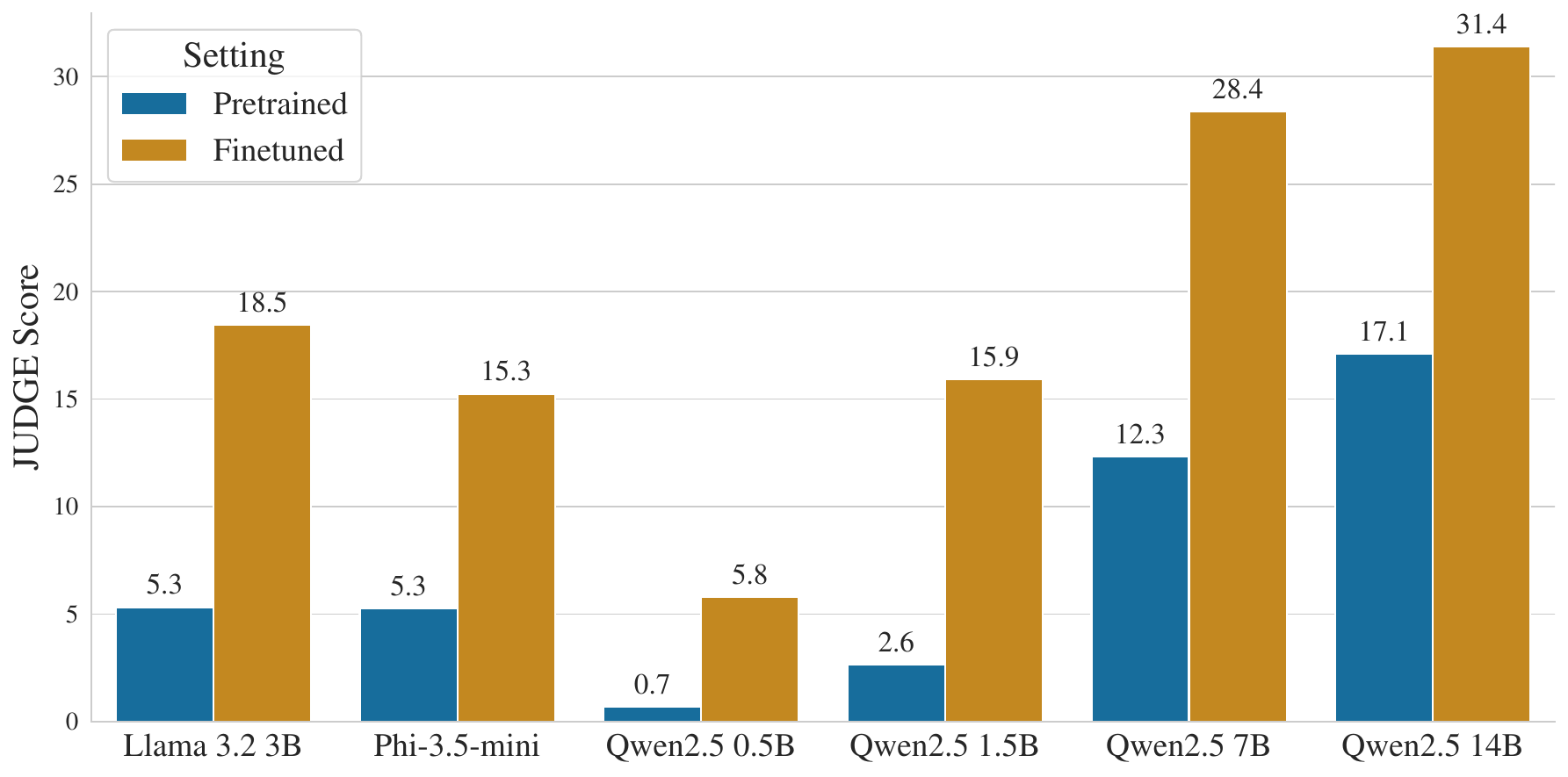}
    \caption{Comparison of the JUDGE scores between pre-trained and fine-tuned models on the test split of \gls{SLDS}. Fine-tuned models outperform the pre-trained models by a large margin.}
    \label{fig:off-the-shelf-vs-finetuned}
\end{figure}

To investigate how well smaller pre-trained models perform in a zero-shot setting, we compare them with their fine-tuned counterparts in \Cref{fig:off-the-shelf-vs-finetuned}. We observe a large performance gap in terms of the JUDGE score between the two settings, highlighting the benefits of fine-tuning on \gls{SLDS}. This pattern is also present in the other metrics, as shown in \Cref{tab:app:off-the-shelf-vs-finetuned}.

\begin{table}[htb]
    \resizebox{\columnwidth}{!}{%
        \begin{tabular}{@{}llrrr@{}}
            \toprule
            \textbf{Model} & \textbf{Metric} & \multicolumn{1}{l}{\textbf{Pretrained}} & \multicolumn{1}{l}{\textbf{Finetuned}} & \multicolumn{1}{l}{\textbf{Delta}} \\ \midrule
            Llama 3.2 3B   & BERTScore       & -9.41 ± 3.51                            & 15.2 ± 4.40                            & +24.61                             \\
            Llama 3.2 3B   & BLEU            & 2.81 ± 0.24                             & 21.89 ± 0.42                           & +19.08                             \\
            Llama 3.2 3B   & ROUGE-1         & 17.36 ± 2.26                            & 31.89 ± 2.34                           & +14.53                             \\
            Llama 3.2 3B   & ROUGE-2         & 5.02 ± 0.91                             & 14.87 ± 1.61                           & +9.85                              \\
            Llama 3.2 3B   & ROUGE-L         & 9.89 ± 1.21                             & 22.49 ± 1.60                           & +12.60                             \\
            Llama 3.2 3B   & JUDGE           & 5.33 ± 1.65                             & 18.47 ± 2.99                           & +13.14                             \\ \midrule
            Phi-3.5-mini   & BERTScore       & -17.1 ± 4.02                            & 11.24 ± 3.82                           & +28.34                             \\
            Phi-3.5-mini   & BLEU            & 2.36 ± 0.17                             & 34.84 ± 0.41                           & +32.48                             \\
            Phi-3.5-mini   & ROUGE-1         & 16.18 ± 2.42                            & 31.2 ± 2.08                            & +15.02                             \\
            Phi-3.5-mini   & ROUGE-2         & 4.25 ± 0.77                             & 14.11 ± 1.27                           & +9.86                              \\
            Phi-3.5-mini   & ROUGE-L         & 8.5 ± 1.16                              & 20.96 ± 1.35                           & +12.46                             \\
            Phi-3.5-mini   & JUDGE           & 5.28 ± 1.72                             & 15.25 ± 2.32                           & +9.97                              \\ \midrule
            Qwen2.5 0.5B   & BERTScore       & -16.43 ± 3.28                           & -1.37 ± 3.85                           & +15.06                             \\
            Qwen2.5 0.5B   & BLEU            & 4.67 ± 0.16                             & 32.2 ± 0.35                            & +27.53                             \\
            Qwen2.5 0.5B   & ROUGE-1         & 11.75 ± 1.67                            & 23.87 ± 1.68                           & +12.12                             \\
            Qwen2.5 0.5B   & ROUGE-2         & 2.19 ± 0.43                             & 9.46 ± 0.94                            & +7.27                              \\
            Qwen2.5 0.5B   & ROUGE-L         & 7.33 ± 0.93                             & 17.37 ± 1.09                           & +10.04                             \\
            Qwen2.5 0.5B   & JUDGE           & 0.68 ± 0.35                             & 5.8 ± 1.26                             & +5.12                              \\ \midrule
            Qwen2.5 1.5B   & BERTScore       & -23.85 ± 4.90                           & 19.81 ± 2.72                           & +43.66                             \\
            Qwen2.5 1.5B   & BLEU            & 3.9 ± 0.18                              & 36.79 ± 0.34                           & +32.89                             \\
            Qwen2.5 1.5B   & ROUGE-1         & 15.62 ± 1.92                            & 33.03 ± 1.73                           & +17.41                             \\
            Qwen2.5 1.5B   & ROUGE-2         & 3.58 ± 0.65                             & 14.14 ± 1.08                           & +10.56                             \\
            Qwen2.5 1.5B   & ROUGE-L         & 9.27 ± 0.99                             & 22.67 ± 1.13                           & +13.40                             \\
            Qwen2.5 1.5B   & JUDGE           & 2.64 ± 0.98                             & 15.92 ± 2.27                           & +13.28                             \\ \midrule
            Qwen2.5 3B     & BERTScore       & -7.82 ± 3.28                            & 23.23 ± 2.80                           & +31.05                             \\
            Qwen2.5 3B     & BLEU            & 5.55 ± 0.20                             & 38.42 ± 0.34                           & +32.87                             \\
            Qwen2.5 3B     & ROUGE-1         & 20.18 ± 2.03                            & 35.18 ± 1.79                           & +15.00                             \\
            Qwen2.5 3B     & ROUGE-2         & 4.96 ± 0.83                             & 15.66 ± 1.23                           & +10.70                             \\
            Qwen2.5 3B     & ROUGE-L         & 11.25 ± 1.01                            & 24.1 ± 1.17                            & +12.85                             \\
            Qwen2.5 3B     & JUDGE           & 6.18 ± 1.78                             & 20.31 ± 2.66                           & +14.13                             \\ \midrule
            Qwen2.5 7B     & BERTScore       & -11.41 ± 5.52                           & 29.59 ± 1.97                           & +41.00                             \\
            Qwen2.5 7B     & BLEU            & 4.03 ± 0.25                             & 41.4 ± 0.34                            & +37.37                             \\
            Qwen2.5 7B     & ROUGE-1         & 20.24 ± 2.50                            & 39.24 ± 1.59                           & +19.00                             \\
            Qwen2.5 7B     & ROUGE-2         & 6.31 ± 1.03                             & 18.26 ± 1.25                           & +11.95                             \\
            Qwen2.5 7B     & ROUGE-L         & 11.42 ± 1.33                            & 26.44 ± 1.15                           & +15.02                             \\
            Qwen2.5 7B     & JUDGE           & 12.34 ± 2.39                            & 28.37 ± 3.07                           & +16.03                             \\ \midrule
            Qwen2.5 14B    & BERTScore       & -19.02 ± 7.35                           & 32.48 ± 1.98                           & +51.50                             \\
            Qwen2.5 14B    & BLEU            & 4.85 ± 0.28                             & 41.8 ± 0.37                            & +36.95                             \\
            Qwen2.5 14B    & ROUGE-1         & 20.02 ± 2.76                            & 40.04 ± 1.74                           & +20.02                             \\
            Qwen2.5 14B    & ROUGE-2         & 7.21 ± 1.17                             & 19.99 ± 1.41                           & +12.78                             \\
            Qwen2.5 14B    & ROUGE-L         & 11.04 ± 1.48                            & 28.0 ± 1.28                            & +16.96                             \\
            Qwen2.5 14B    & JUDGE           & 17.11 ± 2.98                            & 31.38 ± 3.19                           & +14.27                             \\ \bottomrule
        \end{tabular}%
    }
    \caption{Comparison of pre-trained and fine-tuned models on the \gls{SLDS} test split. All metrics show a substantial increase after fine-tuning.}
    \label{tab:app:off-the-shelf-vs-finetuned}
\end{table}

\section{Distribution of Judgment Scores}
\label[appendix]{app:judge-human-score-distributions}

\begin{figure*}[h]
    \centering
    \begin{subfigure}{0.49\textwidth}
        \centering
        \includegraphics[width=\textwidth]{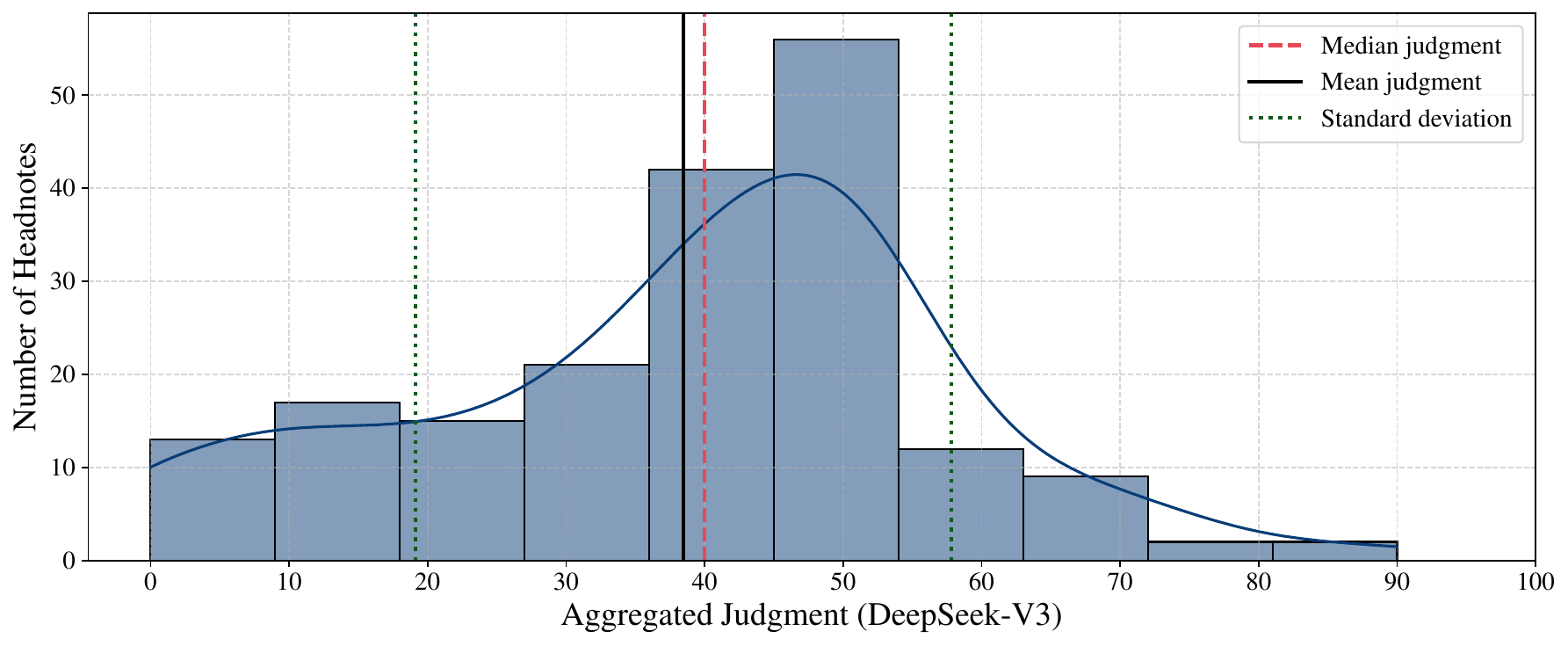}
        \caption{LLM-as-a-Judge}
        \label{fig:judge_and_human_score_distributions:judge}
    \end{subfigure}
    \begin{subfigure}{0.49\textwidth}
        \centering
        \includegraphics[width=\textwidth]{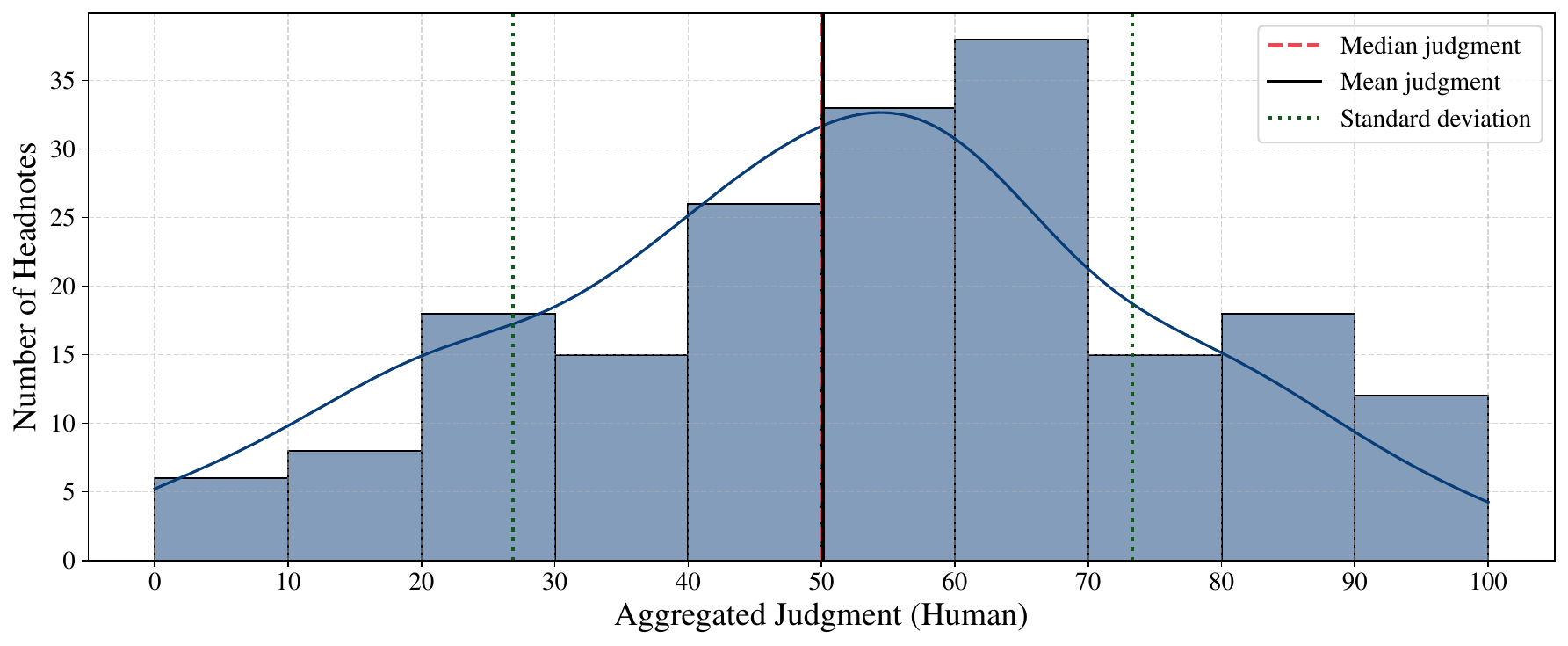}
        \caption{Human-as-a-Judge}
        \label{fig:judge_and_human_score_distributions:human}
    \end{subfigure}
    \vspace{-1ex}
    \caption{Distributions of (a) the scores generated by DeepSeek-V3 and (b) the scores assigned by two lawyers. The scores are aggregates of the individual scores per evaluation category, ranging from 0 to 100. The scores issued by the lawyers are slightly higher than the ones assigned by DeepSeek-V3.}
    \vspace{-2ex}
    \label{fig:judge_and_human_score_distributions}
\end{figure*}

\Cref{fig:judge_and_human_score_distributions} provides an overview of the scores that were assigned by the \gls{LLM} judge (left) and the human judges (right).

\section{Contextualized Expert Evaluation of LLM-Generated Headnotes}
\label[appendix]{app:contextualized-evaluation}

In \Crefrange{fig:app:contextualized-evaluation-sample-1}{fig:app:contextualized-evaluation-sample-6}, we present summaries of the expert commentary provided by our legal expert as part of the contextualized evaluation. The expert reviewed each case with access to the full decision, the official (gold) headnote, and the headnotes generated by different models, without being informed of the model identities. To save space, we do not reproduce the full decisions here; however, they remain accessible via their sample IDs in the SLDS dataset on Hugging Face. We present English translations of the official headnote as well as translations of the model-generated versions from Claude 3.5 Sonnet, DeepSeek R1, and our fine-tuned Qwen2.5 14B.

\begin{figure*}[tb]
    \vspace{-2ex}
    \begin{contextualizedEvaluation}
        {60465}
        {
            Art. 10 and 13 ECHR; Art. 16, 29a, 35 and 93 para. 3 Cst.; Art. 2 let. cbis, Art. 5a, 25 para. 3 let. b, Art. 83 para. 1 let. a and Art. 93 para. 1 and 95 para. 1 RTVA; Art. 28 ff. CC; Art. 1, 3, 5 para. 4 and Art. 18 of the SRG Concession; Deletion of a user comment on Instagram by the SRG in other journalistic offerings (OJO); Legal recourse. The SRG is bound by fundamental rights in its other journalistic offerings (OJO); this applies - because of the close content-related connection between its editorial contribution and the user comments thereto - also insofar as it deletes such comments based on its "Netiquette" (E. 2). Since, due to lack of effectiveness, neither civil, criminal nor supervisory legal recourse in this context meets the requirements of Art. 29a Cst., the administrative legal recourse via the SRG Ombuds Office, which has a mediating function, is to be opened to the Independent Complaints Authority for Radio and Television (ICA) (E. 3). The specifications in the "Netiquette" are to be handled in individual cases analogous to the jurisprudence on the fundamental rights commitment of the SRG in the advertising sector (BGE 139 I 306) (E. 4).
        }
        {
            Art. 29a BV, Art. 35 para. 2 BV, Art. 83 para. 1 lit. a RTVG; Fundamental rights obligation of SRG in the deletion of social media comments and legal process. The SRG is bound by fundamental rights when deleting user comments on editorial contributions in the remaining journalistic offerings (RJO) on social media platforms. The deletion of a comment constitutes an evaluative editorial act that interferes with the freedom of expression of the commenting person. The guarantee of legal recourse (Art. 29a BV) requires effective judicial protection. Neither the civil and criminal legal route nor the supervisory procedure at BAKOM meets these constitutional requirements. The UBI is responsible for assessing the legality of comment deletions, and complaints must be submitted there after an unsuccessful ombudsman procedure (E. 2-4).
        }
        {
            The system generates only a limited number of references compared to the original headnote and excerpt. It identifies the core of the decision quite accurately and uses terminology similar to both the excerpt and the original headnote. While the part on the “deletion of comments” is correct, the addition of “interference with the personal freedom” is problematic, as it is not mentioned in the decision or the original headnote. Regarding “legal recourse,” the output partially contradicts the main judgment: whereas the decision affirms that a civil law remedy can meet the requirements, the headnote suggests otherwise. However, the part on judicial protection is accurate. The role of the UBI is summarized precisely.
        }
        {
            Art. 35 Cst, Art. 29a Cst, Art. 83 LRTV; Obligation of the SRG to adhere to fundamental rights when deleting user-generated comments in the rest of the journalistic offer. The Swiss Broadcasting Corporation (SRG) is bound to fundamental rights when moderating comments in its social media channels within the framework of the rest of the journalistic offer (üjO), as it acts in the exercise of a public task (Art. 35 para. 2 Cst). The close factual connection between editorial contributions and user comments establishes a functional unit, which obliges the SRG to balance conflicting interests in conformity with fundamental rights (para. 2.2.3, 2.3.1). The Independent Complaints Authority for Radio and Television (ICA) is competent under Art. 83 para. 1 letter a RTVG to decide on complaints against the deletion of comments, as these are to be qualified as editorial factual acts in the sense of Art. 2 letter cbis RTVG (para. 3.3.1, 3.3.4). The civil or supervisory legal process does not meet the requirements of the guarantee of legal recourse (Art. 29a Cst) and effective legal protection (Art. 13 ECHR) because it does not allow for individual fundamental rights control (para. 3.2.2, 3.2.3). The SRG must justify the deletion of comments based on objective criteria, whereby the "netiquette" is to be interpreted in the light of freedom of expression (Art. 16 Cst) and may only intervene if there are significant reasons (e.g., discrimination, glorification of violence) (para. 4.1, 4.2). The ICA must examine in individual cases whether the SRG has respected its obligation to fundamental rights.
        }
        {
            The generated headnote includes fewer references than the original headnote and excerpt. It states the context of the case precisely. However, it focuses more on the role of the SRG and includes the appellant's claim under Art. 35 para. 2. Since such claims are part of a party's pleading, they should not be included in a headnote. This is why the original headnote does not mention them. Overall, the output reads more like a summary than a proper headnote.
        }
        {
            Art. 35 para. 2, Art. 16 para. 2, Art. 29a Cst.; Art. 10 ECHR; Art. 2 let. cbis, Art. 83 para. 1 let. a, Art. 86 para. 1, Art. 94 et seq. RTVA; Comment deletion by SRG in the remaining journalistic offering (RJO). The comment function on social media contributions within the remaining journalistic offering (RJO) is closely linked to the concessionary tasks assigned to SRG in this regard and financed by radio and television fees in the program area (E. 2.2.2). Therefore, SRG is also bound by fundamental rights in this context (E. 2.2.3). The appeals body (UBI) is competent to assess whether SRG impermissibly infringed on the freedom of expression of the comment author by deleting a comment in the remaining journalistic offering (RJO) (E. 3 and 4).
        }
        {
            This is the only output that includes a reference to the ECHR along with other legal provisions. Although not mentioned in the original headnote, this reference is legally important. The roles of the named bodies are stated precisely without unnecessary detail, which is appropriate for a headnote.
        }
    \end{contextualizedEvaluation}
    \vspace{-2ex}
    \begin{adjustwidth}{-1cm}{-1cm}
        \caption{Summarized expert commentary on sample 60465. The figure presents English translations of the official (gold) headnote and the model-generated headnotes from Claude 3.5 Sonnet, DeepSeek R1, and our fine-tuned Qwen2.5 14B. The expert reviewed the case with access to the full decision and all headnotes but without knowing model identities.}
        \label[figure]{fig:app:contextualized-evaluation-sample-1}
    \end{adjustwidth}
\end{figure*}

\begin{figure*}[tb]
    \vspace{-2ex}
    \begin{contextualizedEvaluation}
        {
            60487
        }
        {
            Art. 127 para. 1 Cst.; art. 60a LEaux; principle of legality; principle of cost coverage; principle of causality. According to jurisprudence, the principle of cost coverage, together with the principle of equivalence, serves as a substitute for a formal legal basis in determining the amount of a causal contribution (considerations 3.2 and 3.3). The question of whether the principle of cost coverage itself has constitutional status can remain open. In any case, it does not constitute a distinct and autonomous constitutional right from the principle of legality, which should be respected even when the formal law provides sufficiently precise criteria for calculating the contribution (considerations 3.4 and 3.5). If the law on contributions stipulates, in addition to a sufficiently precise calculation basis, compliance with the principle of cost coverage, the latter applies by virtue of the law and not the Constitution (clarification of the case law; consideration 3.6). Art. 60a LEaux does not result in an individual right under federal legislation for cantonal or communal contributions levied in accordance with this provision to comply with the principle of cost coverage (clarification of the case law; considerations 3.7-3.10).
        }
        {
            Art. 60a LEaux; art. 127 para. 1 Cst.; cost coverage principle. The cost coverage principle does not have the status of an independent constitutional right. Its constitutional function is solely to relax the requirements of the formal legal basis when the nature of the contribution allows it. Art. 60a LEaux implements the principle of causality and obliges the cantons to pass on to the responsible parties the costs of construction, operation, maintenance, remediation, and replacement of wastewater disposal facilities used for public purposes, but does not confer upon the responsible parties an individual right under federal law to compliance with the cost coverage principle (considerations 3.5 and 3.8).
        }
        {
            The references are mentioned correctly, matching the original headnote. The output identifies the principles of the case and states them with precision. While the original headnote leaves open the constitutional status of the cost coverage principle, this version clearly states that it is not constitutionally protected, which reflects a direct reading of the judgment. However, in referring to Art. 60a, it incorrectly names the principle of causality instead of the “polluter pays” principle.
        }
        {
            Art. 127 para. 1 Cst., 60a EPA; cost coverage principle. The cost coverage principle does not constitute an autonomous constitutional right but acts as a substitute for the principle of legality when the formal legal basis of a fee does not sufficiently determine its extent. If the law precisely defines the calculation methods of a fee (in this case, water and wastewater connection charges), the cost coverage principle does not apply to constitutional review (recital 3.5). Art. 60a EPA, which enshrines the polluter-pays principle in the financing of purification infrastructure, does not entail a principle of full cost coverage by the fees collected. Its objective is to transfer the burdens from public communities to users, without prohibiting a financial surplus as long as an adequate legal basis exists (recital 3.7.4). The review of cantonal or communal fees concerning the cost coverage principle falls, in the absence of alleged violation of federal law or cantonal constitutional rights, within the sovereign discretion of local authorities. The Federal Court conducts only a limited arbitrary review (art. 9 Cst.) when formal legality is respected (recitals 3.6, 3.9).(Judgment 149 I 305 of the Federal Court)
        }
        {
            The references are correctly identified, along with the theme of the judgment. This output defines the cost coverage principle more clearly than the first one and aligns more closely with the original headnote. It also supports its content with references to the recitals in the judgment, which adds clarity to the headnote.
        }
        {
            Art. 127 paras. 1 and 2, Art. 8 para. 1 and Art. 9 FC; cost coverage principle. The cost coverage is a principle of constitutional law, but it cannot be invoked independently of the principle of legality. When the collection of a contribution is regulated with sufficient precision in formal law, the cost coverage principle does not need to be examined from the perspective of the Federal Constitution (para. 3.5). Art. 60a EPA does not contain a cost coverage principle (para. 3.7).
        }
        {
            The output correctly identifies both the reference and the theme of the judgment. Its language differs from the other two and presents the cost coverage principle as constitutionally grounded, based on an explanation from the judgment itself. While the original headnote left this question open, this version takes a clear position by engaging more deeply with the text. The use of recitals further supports and clarifies the headnote content.
        }
    \end{contextualizedEvaluation}
    \vspace{-2ex}
    \begin{adjustwidth}{-1cm}{-1cm}
        \caption{Summarized expert commentary on sample 60487. The figure presents English translations of the official (gold) headnote and the model-generated headnotes from Claude 3.5 Sonnet, DeepSeek R1, and our fine-tuned Qwen2.5 14B. The expert reviewed the case with access to the full decision and all headnotes but without knowing model identities.}
        \label[figure]{fig:app:contextualized-evaluation-sample-2}
    \end{adjustwidth}
\end{figure*}

\begin{figure*}[tb]
    \vspace{-2ex}
    \begin{contextualizedEvaluation}
        {
            60490
        }
        {
            Art. 83 let. a LTF; art. 32 para. 1 let. a LTAF; art. 8 and 13 ECHR; violation of the principle of specialty in international administrative assistance; standing to appeal against acts of the Federal Council concerning foreign relations; positive obligations of the State. An intervention by the Federal Council with France on the grounds that the authorities of that State allegedly violated the principle of specialty pertains to foreign relations within the meaning of Art. 32 para. 1 let. a LTAF (paras. 4 and 5). However, access to the courts may be opened under the counter-exception of Art. 32 para. 1 let. a in fine LTAF (paras. 6.1 and 6.2). In this case, it is justified, even if the challenged decision originates from the Federal Council in a matter not listed in Art. 33 let. a and b LTAF, to appeal to the Federal Administrative Court before the Federal Supreme Court (para. 6.3). Art. 8 ECHR did not, in this instance, impose a positive obligation on the Federal Council to issue an injunction to France and therefore cannot open access to the courts under Art. 13 ECHR in connection with Art. 8 ECHR (paras. 6.4 and 6.5).
        }
        {
            Art. 32 para. 1 let. a LTAF; art. 8 and 13 ECHR; international administrative assistance in stock market matters; violation of the principle of specialty; foreign relations. The Federal Council's refusal to intervene with a foreign state due to an alleged violation of the principle of specialty constitutes an act related to foreign relations within the meaning of Art. 32 para. 1 let. a LTAF (consideration 5). Art. 8 ECHR does not require the requested state to intervene with the requesting state when the latter violates, after obtaining the information, the conditions under which their transmission had been subordinated. An individual cannot, therefore, rely on Arts. 8 and 13 ECHR to obtain judicial review of the Federal Council's refusal to intervene (consideration 6).
        }
        {
            The output does not include all the references found in the original headnote but identifies the theme of the judgment precisely. It clearly states the source of the decision and supports its content with references to the relevant recitals.
        }
        {
            Art. 32 para. 1 let. a LTAF; art. 83 let. a LTF; art. 8 and 13 ECHR; principle of specialty in matters of international administrative assistance.
            The decision of the Federal Council not to intervene with a foreign state in the case of an alleged violation of the principle of specialty falls under foreign relations and constitutes a political government act, not subject to appeal unless international law guarantees a right to judicial review (paras. 5.2-5.3).

            The transmission of bank data within the framework of international administrative assistance, in accordance with legal requirements (art. 38 para. 2 LBVM), does not generate a positive obligation for the requested state (Switzerland) to monitor a posteriori the respect of the principle of specialty by the requesting state (France), insofar as the alleged violation falls under the foreign territorial jurisdiction and escapes Swiss control in the sense of art. 1 ECHR (paras. 6.5.3-6.5.4).

            Art. 8 ECHR does not require Switzerland to exert diplomatic pressure to correct a non-compliant use of the transmitted data, with the protection of financial data enjoying a wide margin of appreciation and positive obligations being limited to the preliminary phase of transmission (paras. 6.5.1-6.5.2).

            Art. 13 ECHR does not confer an autonomous right to an effective remedy against a refusal of political intervention by the Federal Council, as Switzerland has ensured a preliminary review of the legal conditions for administrative assistance and the individual has legal remedies in the requesting state to contest the violation of the principle of specialty (para. 6.6).Conclusion: The appeal is inadmissible, as decisions related to the management of diplomatic relations fall under the political discretion of the Federal Council and do not require judicial review in the absence of binding international obligation.
        }
        {
            This output identifies all the references and the theme of the judgment, closely matching the original headnote. It supports the main findings with all relevant recitals and summarizes the case contentions precisely. The inclusion of every article and the reference to recitals is accurate, and the language used is clear and appropriate.
        }
        {
            Art. 32 para. 1 let. a APAF; art. 13 ECHR; art. 8 ECHR; art. 184 para. 2 and 182 para. 2 Const.; art. 38 SESTA; art. 6 ECHR; art. 184 para. 3 Const.; art. 86 SCA; art. 32 para. 1 let. a APAF; art. 83 let. a SCA; art. 86 SCA; art. 33 let. a and b APAF; competence of the Federal Administrative Court to rule on complaints regarding international administrative assistance in stock exchange matters. The complaint against the French Republic, for violation of the principle of specialty by the French authorities in relation to international administrative assistance in stock exchange matters, pertains to foreign relations within the meaning of art. 32 para. 1 let. a APAF (consid. 5). The counter-exception of art. 32 para. 1 let. a in fine APAF, namely the right of the interested party to have his complaint judged by a court, is not applicable in this case. Indeed, art. 8 ECHR does not oblige the requested state to intervene with the requesting state when the latter, after having obtained the required information, violates the conditions under which their transmission was subject, especially when such a violation was not foreseeable (consid. 6).
        }
        {
            This output includes some irrelevant references but correctly identifies the theme and states it upfront, unlike the other two versions which do not mention the complaint against the French Republic and instead focus on the role of the Federal Council. The remaining references to the recitals are supported with clear explanations.
        }
    \end{contextualizedEvaluation}
    \vspace{-2ex}
    \begin{adjustwidth}{-1cm}{-1cm}
        \caption{Summarized expert commentary on sample 60490. The figure presents English translations of the official (gold) headnote and the model-generated headnotes from Claude 3.5 Sonnet, DeepSeek R1, and our fine-tuned Qwen2.5 14B. The expert reviewed the case with access to the full decision and all headnotes but without knowing model identities.}
        \label[figure]{fig:app:contextualized-evaluation-sample-3}
    \end{adjustwidth}
\end{figure*}

\begin{figure*}[tb]
    \vspace{-2ex}
    \begin{contextualizedEvaluation}
        {
            60495
        }
        {
            Art. 51 para. 1 Cst.; cantonal popular initiative "for a more democratic urban development in Geneva"; compatibility with superior law; introduction of a municipal fundamental vote in the field of neighborhood planning (plan localisé de quartier [PLQ]). The creation of a municipal fundamental vote in the presence of several PLQ drafts through a cantonal law in the formal sense does not violate Art. 51 para. 1 Cst. (E. 5.1). The introduction of this new political right without a basis in the cantonal constitution itself is compatible with Geneva constitutional law (E. 5.2).
        }
        {
            Art. 51 para. 1 Cst.; Art. 34 para. 2 Cst.; Introduction of new political rights at the municipal level through a formal law.

            The introduction of a new political right at the municipal level (here: municipal vote on several neighborhood development plans) can be done through a formal law and does not necessarily have to be enshrined in the cantonal constitution. Art. 51 para. 1 Cst. does not require that all political rights at the cantonal and municipal level be exhaustively regulated in the formal cantonal constitution (E. 5.1). An extension of political rights by law without a specific constitutional basis is exceptionally permissible when it concerns a clearly defined area and does not contradict the will of the cantonal constitutional legislator (E. 5.2).
        }
        {
            The output correctly identifies the references and the theme. However, its explanation, especially regarding the recital and Art. 51, does not align with the original judgment. The language is also more complex than necessary, which may reduce clarity.
        }
        {
            **Art. 34 para. 2 Cst.; Art. 51 para. 1 Cst.; Art. 5A para. 5 LGZD (GE); Introduction of new political rights at the municipal level through ordinary law.**
            The creation of a new political right at the municipal level - here a binding vote to choose between competing neighborhood development plans (Art. 5A para. 5 LGZD) - does not require a constitutional basis in the cantonal constitution, provided it concerns a clearly defined matter and no systematic provisions of the cantonal constitution oppose it (consid. 5.1-5.3).

            According to Art. 51 para. 1 Cst., the cantons are only required to have a democratic constitution with an elected parliament and a guarantee of popular rights. The introduction of additional participatory procedures at the legislative level - particularly in the field of municipal spatial planning - does not violate the federal distribution of competences nor the cantonal constitutional law, provided the cantonal constitution does not foresee a *numerus clausus* of popular rights and the regulation is limited and system-consistent (consid. 5.2.1-5.2.2).

            The qualification of a norm as constitutionally relevant in the material sense does not create an obligation to enshrine it in the formal cantonal constitution. Particularly for municipal political rights, statutory regulation is sufficient, especially since public participation in spatial planning is provided for by federal and cantonal law (Art. 4 RPG, Art. 134 Cst./GE; consid. 5.2.2).

            The partial invalidation of the initiative by the cantonal authority thus violates the constitutional right of initiative (Art. 34 para. 2 Cst.), as the contested provision is compatible with superior law.

            *(BGE 149 I 33; references to ATF 104 Ia 343, 142 I 216; HANGARTNER/KLEY, GRISEL)*
        }
        {
            This output includes additional references and demonstrates an understanding of the judgment’s context. However, it elaborates on aspects that are not relevant to a headnote. It does not rely on exact references and adds excessive detail that goes beyond what is appropriate for this format.
        }
        {
            Art. 51 para. 1 Cst; Art. 34 para. 2 Cst; Art. 44 para. 1 and Art. 60 para. 4 Cst/GE; Inclusion of a municipal voting right in a formal cantonal legislative proposal. The introduction of a municipal voting right on several neighborhood plan projects can be incorporated into a formal cantonal legislative proposal in the area of spatial planning without violating Art. 51 para. 1 Cst (E. 5.1). The constitutional law of the canton of Geneva does not contain a general norm stipulating that material constitutional norms must necessarily be included in the constitution in the formal sense (E. 5.2).
        }
        {
            The output identifies the references but includes some inaccuracies, such as citing Art. 44 instead of p. 40 n. 44. Despite this, it is more concise than the other outputs and correctly captures the theme of the case.
        }
    \end{contextualizedEvaluation}
    \vspace{-2ex}
    \begin{adjustwidth}{-1cm}{-1cm}
        \caption{Summarized expert commentary on sample 60495. The figure presents English translations of the official (gold) headnote and the model-generated headnotes from Claude 3.5 Sonnet, DeepSeek R1, and our fine-tuned Qwen2.5 14B. The expert reviewed the case with access to the full decision and all headnotes but without knowing model identities.}
        \label[figure]{fig:app:contextualized-evaluation-sample-4}
    \end{adjustwidth}
\end{figure*}

\begin{figure*}[tb]
    \vspace{-2ex}
    \begin{contextualizedEvaluation}
        {
            60555
        }
        {
            Art. 17, 29 para. 1 and 93 para. 2 Cst; Art. 4 RTVG; television and radio broadcasts about deaths in Ticino nursing homes during the first wave of the pandemic; investigative journalism. Scope and content of media freedom protected by Art. 17 and 93 para. 2 Cst (E. 3.1). Minimum requirements according to Art. 4 para. 1 and 2 RTVG for an editorial program with informational content (E. 3.2 and 3.3). Scope and limits of investigative journalism (E. 3.4). Review authority of the UBI (E. 4.3 and 4.4). The contested television and radio broadcasts in the present case did not violate Art. 4 RTVG (E. 5).
        }
        {
            Art. 4 para. 2 RTVG; Art. 97 para. 2 lit. a RTVG; Requirement of impartiality in radio and television broadcasts; Limits of the UBI's examination authority.

            The UBI exclusively examines the content of already broadcast editorial programs and not their creation process. The examination is conducted in the public interest and not for the protection of the personality rights of those involved (E. 4.4.1). In journalistic research and interviews, inadequate information provided to those involved about the context and the dissemination channel is only relevant if it leads to a violation of Art. 4 para. 2 RTVG, i.e., if it impairs the audience's ability to form their own opinion (E. 4.4.2). Investigative journalism with a critical attitude is also permissible as long as the program as a whole does not appear manipulative and allows the audience to form their own opinion (E. 3.4).
        }
        {
            The output provides insufficient references and fails to identify the primary theme of the case. Instead, it refers to several other aspects that are part of the judgment but are less central to the decision.
        }
        {
            Art. 17 and 93 Swiss Constitution; Art. 4 of the Swiss Federal Act on Radio and Television (RTVG); journalistic duty of care in reporting on COVID-19 deaths in nursing homes.
            The Federal Supreme Court confirms that critical reporting on pandemic management in nursing homes does not violate program law requirements, as long as the principles of factual accuracy and transparency are maintained (E. 3.3, 5.2.5). The freedom of the media according to Art. 17 of the Swiss Constitution also includes investigative journalism, provided that the presentation is not manipulative and allows the audience to form independent judgments (E. 3.4, 5.2.4). There is no violation of Art. 4 para. 2 of the RTVG if emotive elements (e.g., music, interview conduct) can be considered appropriate in the context of a complex and already publicly discussed topic, and the responsible parties involved were able to comprehensively express their views (E. 5.1.3, 5.2.6). The allegation of a formal denial of justice (Art. 29 of the Swiss Constitution) fails because the supervisory complaint with the Independent Complaints Authority for Radio and Television (AIRR) according to Art. 97 RTVG only examines the final program content and not internal editorial agreements (E. 4.4.1). The dignity of those affected (Art. 7 of the Swiss Constitution) is preserved, as the reporting did not equate to a "trial by media" and did not distort the core statements of the interviewees (E. 5.2.5). The jurisprudence on balance (cf. BGE 137 I 340) remains unaffected: A quantitative equal distribution of perspectives is not required as long as divergent viewpoints remain recognizable (E. 3.3, 5.2.4).
        }
        {
            The references are correctly identified, and the primary theme of the case is captured more accurately than in the first output. However, the headnote includes too much detail, which makes it overly bulky for its purpose.
        }
        {
            Art. 4 para. 2 and Art. 97 para. 2 lit. a RTVG; Art. 29 para. 1 BV; Transmission of information about the conditions in a nursing home during the first pandemic wave; Examination of the correctness of the presentation of facts; Procedure. The appellate instance can only resort to the subsequent review of the content of broadcasts to examine compliance with the provisions on the content of editorial broadcasts. The preparations and discussions between journalists and those affected, which preceded, fall outside its scope of examination (E. 4). The review of compliance with the provisions on the content of editorial broadcasts is not aimed at protecting the personal rights of those who participated in the broadcast. The question of whether the broadcast, aired as part of an "investigation," presents the position established by the journalists without depicting it as an absolute truth must be assessed based on the specific circumstances. In this context, the severity of the issue and the significance of the criticism must also be considered. In the present matter, the broadcast did not present the journalists' established position as the only truth, and those affected, who participated in the broadcast, were able to engage with the accusations made against them. The broadcast was therefore not manipulative. The principle of correctness of fact presentation according to Art. 4 para. 2 RTVG has not been violated (E. 5).
        }
        {
            The output correctly identifies the references and the theme at the beginning. However, it introduces the principle of correctness, which does not appear in the original excerpt of the judgment.
        }
    \end{contextualizedEvaluation}
    \vspace{-2ex}
    \begin{adjustwidth}{-1cm}{-1cm}
        \caption{Summarized expert commentary on sample 60555. The figure presents English translations of the official (gold) headnote and the model-generated headnotes from Claude 3.5 Sonnet, DeepSeek R1, and our fine-tuned Qwen2.5 14B. The expert reviewed the case with access to the full decision and all headnotes but without knowing model identities.}
        \label[figure]{fig:app:contextualized-evaluation-sample-5}
    \end{adjustwidth}
\end{figure*}

\begin{figure*}[tb]
    \vspace{-2ex}
    \begin{contextualizedEvaluation}
        {
            60782
        }
        {
            Art. 276 and 285 CC; child support; distribution of excess in the case of children of unmarried parents. Summary of the jurisprudence on the concrete two-stage method with distribution of the excess (para. 2.4-2.6). If only one parent is required to pay the support contribution, the calculation is made between this parent and the beneficiary children of the contribution. An excess that remains after covering the family law subsistence minimum is distributed (in the ratio of two to one) between this parent and the children (para. 2.7).
        }
        {
            Art. 285 para. 1 CC; child maintenance; distribution of surplus between unmarried parents. In the case of unmarried parents, the calculation of the maintenance contribution according to the two-phase concrete method with surplus distribution occurs only between the obligated parent and the child. A virtual share of the surplus cannot be attributed to the unmarried custodial parent who is not entitled to maintenance and is not authorized to actually participate in the surplus of the other parent. The distribution of the surplus must be limited to the persons actually involved in the maintenance relationship (consideration 2.7).
        }
        {
            The output does not identify all relevant references but correctly states the theme of the case. It presents the content concisely and with appropriate reference, demonstrating a good understanding of the judgment.
        }
        {
            Art. 285 CC; concrete two-phase method with excess distribution. In calculating child support for children of unmarried parents, for the purpose of distributing the excess according to the principle of "big and small heads," an additional "big head" for the parent not obliged to make monetary contributions is not to be considered virtually (cons. 2.7). The excess is to be distributed exclusively between the child ("small head") and the obligated parent ("big head"), in the proportion of 1/3 and 2/3, without fictitious deductions in favor of the unmarried custodial parent, who is not entitled to an excess. This ensures a fair assessment of the obligated parent's ability to contribute, in accordance with the dynamic nature of maintenance under Art. 285 CC, avoiding artificial reductions to the detriment of the child's rights (reference to BGE 147 III 265). The claimed automatic equalization with the children of married parents, through the insertion of an additional "big head," is contrary to the principles of the concrete method and the rationale of the child maintenance revision.
        }
        {
            The output correctly identifies the reference and shows an understanding of the case at the outset. However, it misrepresents the core of the calculation discussed in the judgment and uses overly complex language. The judgment outlines a two-step calculation for maintenance, which is not clearly reflected here. Legally, this makes the headnote unnecessarily complicated and suggests a potential misunderstanding of the decision.
        }
        {
            Art. 276 et seq. CC; calculation of the maintenance contribution for the education of children of unmarried parents. If, in applying the profit-sharing calculation method, a profit occurs during the minority of the child, it must be shared between the parent obliged to contribute and the child according to the principle of "big and small head," or according to any other possible justified distribution of the profit for the specific case (consid. 2.1-2.5). If the parents are not married and the child is under the care of one of them, the profit must be shared between the parent obliged to contribute and the child, without considering the non-contributing parent as a "big head" (consid. 2.6 and 2.7).
        }
        {
            The output only partially identifies the relevant reference. It focuses on child support in relation to education, while the judgment addresses child support in the context of caregiving for children of unmarried parents. The language used is more complex than necessary and does not reflect the clarity of the original judgment.
        }
    \end{contextualizedEvaluation}
    \vspace{-2ex}
    \begin{adjustwidth}{-1cm}{-1cm}
        \caption{Summarized expert commentary on sample 60782. The figure presents English translations of the official (gold) headnote and the model-generated headnotes from Claude 3.5 Sonnet, DeepSeek R1, and our fine-tuned Qwen2.5 14B. The expert reviewed the case with access to the full decision and all headnotes but without knowing model identities.}
        \label[figure]{fig:app:contextualized-evaluation-sample-6}
    \end{adjustwidth}
\end{figure*}

%% file: examples/judge/61194/generated_headnote.tex
**Art. 9, 13, 15, 20 und 10bis GebV SchKG; Gebühren und Entschädigungen im Betreibungsverfahren.**  
1. Für die Zustellung eines Zahlungsbefehls können neben der Gebühr nach Art. 16 Abs. 1 GebV SchKG Auslagen für Posttaxen (Art. 13 Abs. 1 GebV SchKG) verrechnet werden. Ein erfolgloser Zustellversuch löst jedoch erst ab dem zweiten Versuch eine zusätzliche Gebühr nach Art. 16 Abs. 3 GebV SchKG aus (E. 3.2.1–3.2.2).  
2. **Abholungseinladungen** für Zahlungsbefehle stellen keine gesetzlich vorgeschriebenen Amtshandlungen dar. Für sie dürfen weder Gebühren nach Art. 9 GebV SchKG noch Auslagen nach Art. 13 GebV SchKG erhoben werden, selbst wenn Art. 10bis GebV SchKG dies seit 2022 vorsieht, da die Verordnung hierfür keine hinreichende gesetzliche Grundlage bietet (E. 3.2.3).  
3. **Pfändungsankündigungen** sind vom Pfändungsvollzug zu trennen und können separat mit einer Gebühr nach Art. 9 Abs. 1 lit. a GebV SchKG sowie Auslagen für eingeschriebene Zustellung belastet werden. Ein zusätzlicher Versand per A-Post ist jedoch mangels gesetzlicher Grundlage nicht kostenpflichtig (E. 3.3.1–3.3.2).  
4. **Verlustscheine** (Art. 115 SchKG) dürfen nur für die Abschrift der Pfändungsurkunde (Art. 24 GebV SchKG) und deren eingeschriebene Zustellung belastet werden. Die Zustellung per A-Post an den Schuldner ist unzulässig und damit nicht erstattungsfähig (E. 3.4).  
5. **Wegentschädigungen** nach Art. 14 GebV SchKG setzen voraus, dass das Betreibungsamt mehrere Verrichtungen gemäss Art. 15 GebV SchKG berücksichtigt. Die Nichtbeachtung entsprechender Parteivorbringen verletzt das rechtliche Gehör (Art. 29 Abs. 2 BV) und führt zur Zurückweisung des Entscheids (E. 3.5.1).  
Die Gebührenverordnung ist restriktiv auszulegen: Kosten dürfen nur für gesetzlich vorgeschriebene Amtshandlungen erhoben werden, wobei die Praxis der Betreibungsämter keine eigenständige Rechtsgrundlage schafft (E. 3.2.3, 3.3.2).

%% file: examples/judge/61194/generated_headnote_translated.tex
**Art. 9, 13, 15, 20, and 10bis GebV SchKG; Fees and compensations in debt enforcement proceedings.**
In addition to the fee under Art. 16 para. 1 GebV SchKG, expenses for postal charges (Art. 13 para. 1 GebV SchKG) may be charged for the delivery of a payment order. However, an unsuccessful delivery attempt only triggers an additional fee under Art. 16 para. 3 GebV SchKG starting from the second attempt (consid. 3.2.1–3.2.2).
Collection notices for payment orders do not constitute legally prescribed official acts. Therefore, no fees under Art. 9 GebV SchKG or expenses under Art. 13 GebV SchKG may be charged for them, even though Art. 10bis GebV SchKG has provided for this since 2022, as the ordinance lacks a sufficient legal basis for such charges (consid. 3.2.3).
Seizure announcements must be distinguished from the execution of the seizure itself and may be charged separately with a fee under Art. 9 para. 1 lit. a GebV SchKG, along with expenses for registered delivery. However, an additional dispatch by A-Post is not chargeable due to the lack of a legal basis (consid. 3.3.1–3.3.2).
Loss certificates (Art. 115 SchKG) may only be charged for the copy of the seizure record (Art. 24 GebV SchKG) and its registered delivery. Delivery by A-Post to the debtor is not permissible and therefore not reimbursable (consid. 3.4).
Travel compensations under Art. 14 GebV SchKG require that the debt enforcement office considers multiple tasks in accordance with Art. 15 GebV SchKG. Failure to consider relevant submissions by the parties violates the right to be heard (Art. 29 para. 2 BV) and results in the annulment of the decision (consid. 3.5.1).
The fee ordinance must be interpreted restrictively: Costs may only be charged for legally prescribed official acts, and the practices of the debt enforcement offices do not constitute an independent legal basis (consid. 3.2.3, 3.3.2).

%% file: examples/judge/61194/original_headnote.tex
Art. 1, Art. 2, Art. 9 Abs. 1 lit. a, Art. 10bis, Art. 13 Abs. 1, Art. 14, Art. 15 Abs. 1, Art. 16 Abs. 1 und Abs. 3, Art. 20, Art. 24 GebV SchKG; Art. 16, Art. 34, Art. 72 Abs. 1, Art. 90, Art. 112, Art. 114, Art. 115 Abs. 1 SchKG; Kosten von Zahlungsbefehlen, Pfändungsankündigungen und Verlustscheinen.  Allgemeines zu Gebühren und Entschädigungen gemäss GebV SchKG (E. 3.1). 
Kosten für die Zustellung von Zahlungsbefehlen (E. 3.2.1); Gebühr bei einem erfolglosen Zustellversuch (E. 3.2.2) und für eine Abholungseinladung. Art. 10bis GebV SchKG stellt keine genügende gesetzliche Grundlage dar, um für die Einladung zur Abholung eines Zahlungsbefehls Kosten in Rechnung zu stellen (E. 3.2.3). 
Die Kosten für eine Pfändungsankündigung sind nicht in Art. 20 GebV SchKG geregelt (E. 3.3.1). Die Pfändungsankündigung ist nach Art. 34 SchKG zuzustellen. Die Zustellung mit A-Post ist nicht vorgesehen und kann nicht in Rechnung gestellt werden (E. 3.3.2). 
Pfändungsurkunde als Verlustschein (Art. 115 Abs. 1 SchKG). Art. 20 Abs. 1 GebV SchKG bezieht sich nur auf die Abfassung der Pfändungsurkunde für das Amt (Art. 112 SchKG) und nicht auf die Abschriften für den Schuldner und die Gläubiger (Art. 114 SchKG). Gebühren für diese Abschriften (Art. 24 GebV SchKG). Die Abschriften sind nach Art. 34 SchKG zuzustellen. Die Zustellung mit A-Post ist nicht vorgesehen und kann nicht in Rechnung gestellt werden (E. 3.4). 
Wegentschädigungen (Art. 14 und 15 GebV SchKG). Verletzung des rechtlichen Gehörs; Sachverhaltsfeststellung von Amtes wegen (Art. 20a Abs. 2 Ziff. 2 SchKG) und Pflicht der Aufsichtsbehörden, die Anwendung der GebV SchKG zu überwachen (Art. 2 GebV SchKG) (E. 3.5).

%% file: examples/judge/61194/original_headnote_translated.tex
Art. 1, Art. 2, Art. 9 para. 1 let. a, Art. 10bis, Art. 13 para. 1, Art. 14, Art. 15 para. 1, Art. 16 para. 1 and para. 3, Art. 20, Art. 24 GebV SchKG; Art. 16, Art. 34, Art. 72 para. 1, Art. 90, Art. 112, Art. 114, Art. 115 para. 1 SchKG; Costs of payment orders, seizure notices, and loss certificates. General principles on fees and compensations according to the GebV SchKG (consid. 3.1).
Costs for the delivery of payment orders (consid. 3.2.1); fee for an unsuccessful delivery attempt (consid. 3.2.2) and for a pickup invitation. Art. 10bis GebV SchKG does not constitute a sufficient legal basis to charge costs for the invitation to pick up a payment order (consid. 3.2.3).
The costs for a seizure notice are not regulated in Art. 20 GebV SchKG (consid. 3.3.1). The seizure notice must be delivered according to Art. 34 SchKG. Delivery by A-Post is not provided for and cannot be charged (consid. 3.3.2).
Seizure certificate as a loss certificate (Art. 115 para. 1 SchKG). Art. 20 para. 1 GebV SchKG only refers to the drafting of the seizure certificate for the authority (Art. 112 SchKG) and not to the copies for the debtor and creditors (Art. 114 SchKG). Fees for these copies (Art. 24 GebV SchKG). The copies must be delivered according to Art. 34 SchKG. Delivery by A-Post is not provided for and cannot be charged (consid. 3.4).
Travel compensations (Art. 14 and 15 GebV SchKG). Violation of the right to be heard; determination of facts ex officio (Art. 20a para. 2 no. 2 SchKG) and the duty of supervisory authorities to monitor the application of the GebV SchKG (Art. 2 GebV SchKG) (consid. 3.5).

%% file: examples/judge/61194/judge_output.tex
ACCURACY\_FAITHFULNESS: \\
Analysis: The Model-Generated Headnote captures the general legal principles and main points of the Official Headnote but lacks some specific details and precise references. For instance, it omits references to Articles 1, 2, 10bis, 14, 16, 72, 90, 112, and 115 SchKG, which are crucial in the Official Headnote. Additionally, the Model-Generated Headnote introduces new elements like the reference to BGE 138 III 25, which is not present in the Official Headnote. \\
ACCURACY\_FAITHFULNESS\_SCORE: 2 \medskip

COMPLETENESS\_RELEVANCE: \\
Analysis: The Model-Generated Headnote includes several key points from the Official Headnote, such as the treatment of Abholungseinladung, Pfändungsankündigung, Verlustschein, and Wegentschädigung. However, it omits significant details, such as the broader context of general fees and compensations (E. 3.1), the specific mention of Articles 10bis, 14, 15, 16, 72, 90, 112, and 115 SchKG, and the role of supervisory authorities (Art. 2 GebV SchKG). These omissions reduce the completeness and relevance of the Model-Generated Headnote. \\ 
COMPLETENESS\_RELEVANCE\_SCORE: 2  \medskip

CLARITY\_COHERENCE: \\
Analysis: The Model-Generated Headnote is well-organized and coherent, with clear sections addressing different aspects of the case. However, the inclusion of an unnecessary reference to BGE 138 III 25 and the omission of certain key articles and considerations slightly detract from its clarity. The structure is logical, but the lack of completeness affects the overall coherence.  \\
CLARITY\_COHERENCE\_SCORE: 2  \medskip

ARTICLES:  \\
Analysis: The Model-Generated Headnote correctly references Articles 9, 13, 15, 20, 24 GebV SchKG and Articles 34, 114 SchKG. However, it omits several critical articles mentioned in the Official Headnote, such as Articles 1, 2, 10bis, 14, 16, 72, 90, 112, and 115 SchKG. This omission significantly impacts the accuracy and completeness of the legal references.  
ARTICLES\_SCORE: 1  \medskip

CONSIDERATIONS:  \\
Analysis: The Model-Generated Headnote cites considerations E. 3.2.3, E. 3.3.1, E. 3.3.2, E. 3.4, and E. 3.5.1, which are relevant and align with the Official Headnote. However, it misses other critical considerations, such as E. 3.1, E. 3.2.1, E. 3.2.2, and broader references to the role of supervisory authorities. This omission reduces the completeness of the considerations.  \\
CONSIDERATIONS\_SCORE: 2